\documentclass[11pt]{article}

\usepackage[preprint]{acl}

\usepackage{times}
\usepackage{latexsym}

\usepackage[T1]{fontenc}

\usepackage[utf8]{inputenc}

\usepackage{microtype}

\usepackage{inconsolata}

\usepackage{graphicx}

\usepackage{amsmath}
\usepackage{amssymb}
\usepackage{mathtools}
\usepackage{amsthm}
\usepackage[table]{xcolor}
\usepackage{booktabs}
\usepackage{enumitem}
\usepackage{algorithm}
\usepackage{algorithmic}
\usepackage{multirow}
\usepackage{float}
\usepackage{stfloats} 
\usepackage{tikz}
\usetikzlibrary{arrows.meta,shapes.arrows}

\definecolor{gcpTFill}{HTML}{E3EEFA}
\definecolor{gcpTDraw}{HTML}{3A6EA5}
\definecolor{gcpSFill}{HTML}{FCE8CF}
\definecolor{gcpSDraw}{HTML}{B8742A}
\definecolor{gcpGray}{HTML}{EEEEEE}
\definecolor{gcpEFill}{HTML}{F6D3CF}
\definecolor{gcpEDraw}{HTML}{B03A2E}

\AtBeginDocument{%
  \setlength{\abovedisplayskip}{3pt plus 1pt minus 1pt}%
  \setlength{\belowdisplayskip}{3pt plus 1pt minus 1pt}%
  \setlength{\abovedisplayshortskip}{1pt plus 1pt minus 1pt}%
  \setlength{\belowdisplayshortskip}{1pt plus 1pt minus 1pt}%
}

\newtheoremstyle{compactplain}{4pt}{4pt}{\itshape}{}{\bfseries}{.}{.5em}{}
\newtheoremstyle{compactremark}{4pt}{4pt}{}{}{\itshape}{.}{.5em}{}
\theoremstyle{compactplain}
\newtheorem{theorem}{Theorem}
\newtheorem{proposition}[theorem]{Proposition}

\newcommand{\restatedname}{}
\newtheorem*{restatedthm}{\restatedname}
\newenvironment{restated}[1]
  {\renewcommand{\restatedname}{#1}\begin{restatedthm}}
  {\end{restatedthm}}
\theoremstyle{definition}
\newtheorem{definition}{Definition}
\newtheorem{assumption}{Assumption}
\theoremstyle{compactremark}
\newtheorem{remark}{Remark}
\newtheorem*{remark*}{Remark}

\title{Distilling LLM Reasoning into a Graph of Concept Predictors}

\author{Ziyang Yu \and Liang Zhao \\
  Emory University}

\begin{document}
\maketitle
\begin{abstract}
Deploying Large Language Models (LLMs) for discriminative workloads is often
limited by inference latency, compute, and API costs at scale. Active
distillation reduces these costs by querying an LLM oracle to train small
discriminative students, but most pipelines distill only final labels,
discarding intermediate reasoning signals and offering limited diagnostics of
what reasoning is missing and where errors arise. We propose Graph of Concept Predictors (GCP), 
a reasoning-aware active distillation framework in which the teacher's reasoning is 
elicited as a directed acyclic graph of intermediate concepts and mirrored in the student. GCP enhances sample
efficiency through a graph-aware acquisition strategy that weights
per-concept uncertainty, gradient diversity, and coverage by node
centrality. Additionally, it improves
training stability and efficiency by performing targeted sub-module
retraining, which attributes downstream loss to specific concept predictors
and updates only the most influential modules. Experiments on eight NLP
classification benchmarks demonstrate that GCP enhances performance under
limited annotation budgets while yielding more interpretable and controllable
training dynamics. Code is available at
\url{https://anonymous.4open.science/r/GCP-1BB6}.
\end{abstract}
\section{Introduction}\vspace{-1mm}
Large Language Models (LLMs) have increasingly been adopted for
discriminative tasks such as topic classification, sentiment analysis, stance
detection, and clinical risk prediction
~\cite{Brown2020LanguageMA,Cruickshank2023PromptingAF,Rezk2024LLMsFC}. Unlike
traditional classifiers, which must be retrained whenever the label space or
task definition changes, LLMs can generalize to unseen tasks through prompting
~\cite{Kojima2022LargeLM}, making them particularly attractive when task
definitions evolve rapidly. This flexibility has led to widespread deployment
of LLMs as zero-shot or few-shot discriminative models across domains such as
finance, healthcare, and large-scale content moderation
~\cite{Naveed2023ACO}. However, this capability comes at substantial
computational and monetary cost. For example, inference with frontier-scale models can cost over
\$20 per million tokens, and even lightweight models still incur
nontrivial cost when applied to large corpora~\cite{Erol2025CostofPassAE}.
 For data at scale, such as millions of documents
in financial compliance screening and e-commerce
review classification, cumulative inference cost and latency quickly become
prohibitive, which makes direct LLM deployment impractical for accurate,
low cost discriminative workloads~\cite{Bai2024BeyondEA}.

A natural solution to these limitations is to distill LLMs into smaller
models~\cite{Hinton2015DistillingTK}. Early work primarily focused on
distilling large LLMs into smaller language models, preserving generative
capacity while reducing parameter count
~\cite{Sanh2019DistilBERTAD,Jiao2019TinyBERTDB,Sun2020MobileBERTAC}. While
effective, such approaches remain computationally expensive and often fail to
deliver order-of-magnitude savings in inference latency or cost
~\cite{Wan2023EfficientLL}. 
In contrast, more recent research has shown that
LLM behavior can be distilled into task-specific discriminative models (e.g.,
multi-layer perceptrons, linear models, or support vector machines) that
reduce training and inference costs by orders of magnitude while maintaining
strong performance~\cite{Ho2022LargeLM,Bhat2023LargeLM}.
These approaches typically treat the
LLM as an oracle annotator and use active learning to minimize the number of
LLM queries~\cite{Xia2025FromST,Xiang2025PromptALSD}, selecting only the most
informative samples for labeling. The resulting objective is to train a
small discriminative student that approximates the final labels produced by
the LLM, achieving substantial financial and time savings suitable for
real-world deployment~\cite{Shridhar2022DistillingRC}.

A key reason LLMs achieve strong performance on complex discriminative tasks
is their ability to generate and exploit explicit reasoning processes, such
as Chain-of-Thought, Tree-of-Thought, and Graph-of-Thought reasoning
~\cite{Wei2022ChainOT,Yao2023TreeOT,Yao2023BeyondCE}. These paradigms can be
viewed as structured, directed acyclic graphs of intermediate reasoning states
that enable test-time computation scaling and systematic exploration of
decision paths, leading to improved accuracy and robustness
~\cite{Yang2023LargeLM}. A growing body of work has empirically shown that
eliciting intermediate reasoning substantially enhances LLM performance,
particularly on tasks requiring compositional or multi-factor judgment
~\cite{Wei2022ChainOT,Kojima2022LargeLM}. Despite this, existing distillation
pipelines largely discard intermediate reasoning signals, training student
models solely on final outputs in an end-to-end fashion
~\cite{Hinton2015DistillingTK}. This practice results in sparse supervision,
obscures the structure of the decision process, reduces interpretability, and
limits the student's ability to recover the teacher's reasoning competence,
especially when distilled into simple discriminative architectures
~\cite{Ho2022LargeLM}.

\looseness=-1
To approach teacher-level performance, a small discriminative model must
learn not only what prediction to make, but also how to arrive at it
~\cite{Wei2022ChainOT}. This motivates distilling the reasoning process itself
rather than only final labels. 
However, incorporating reasoning supervision
fundamentally complicates the active distillation pipeline. For sample
selection, reasoning introduces structured, multi-step dependencies:
different examples may expose weaknesses in different parts of the reasoning
process, making flat uncertainty or diversity criteria over final predictions
insufficient~\cite{Wei2022ChainOT,Yao2023TreeOT}. 
Selecting samples that
improve reasoning fidelity requires estimating, before the teacher is
queried, where in the reasoning the student is likely to err. For model retraining, reasoning supervision localizes errors
to specific intermediate components rather than the final output alone,
raising the challenge of attributing downstream loss to internal decision
modules and updating them efficiently without retraining the entire model
~\cite{Huang2022TowardsRI}. These challenges affect both the efficiency and
stability of active distillation when reasoning is taken seriously.

\begin{figure}[t!]
\centering
\begin{tikzpicture}[
    x=1cm, y=1cm,
    font=\scriptsize,
    box/.style={rounded corners=2.5pt, align=flush center, line width=0.6pt,
                text width=1.55cm, minimum height=0.62cm, inner sep=1.5pt},
    tnode/.style={box, draw=gcpTDraw, fill=gcpTFill},
    snode/.style={box, draw=gcpSDraw, fill=gcpSFill},
    inedge/.style={-{Stealth[length=3.5pt,width=3pt]}, line width=0.6pt, draw=gray!70},
    tedge/.style={-{Stealth[length=3.5pt,width=3pt]}, line width=0.7pt, draw=gcpTDraw},
    sedge/.style={-{Stealth[length=3.5pt,width=3pt]}, line width=0.7pt, draw=gcpSDraw},
    note/.style={fill=gcpGray, rounded corners=2pt, inner sep=2.5pt, align=center},
  ]
  \def\xL{1.85}  \def\xR{5.85}  \def\dx{0.92}
  \def\yT{0}     \def\yX{-0.62}
  \def\yA{-1.5}  \def\yB{-2.45} \def\yC{-3.4} \def\yF{-4.25}

  \node[font=\footnotesize\bfseries, text=gcpTDraw] at (\xL,\yT) {Teacher LLM};
  \node[font=\footnotesize\bfseries, text=gcpSDraw] at (\xR,\yT) {Student (GCP)};
  \node[single arrow, draw=gray!60, fill=gcpGray, minimum height=1.1cm,
        single arrow head extend=2pt, inner sep=1.2pt]
    (mirror) at (3.85,\yT) {mirror};
  \draw[line width=0.8pt, gray!70] (mirror.south) -- (3.85,\yF-0.4);

  \node[draw=gray!60, fill=white, rounded corners=2.5pt, inner sep=2.5pt,
        line width=0.5pt, minimum width=7.5cm] (x) at (3.85,\yX)
    {\textbf{Input $x$} (Yelp review):\ \emph{``Great food, but we waited
     an hour.''}};

  \node[tnode] (t1) at (\xL-\dx,\yA) {$R_1$: food\\ good};
  \node[tnode] (t2) at (\xL+\dx,\yA) {$R_2$: service\\ slow};
  \node[tnode] (t3) at (\xL,\yB)     {$R_3$: visit\\ soured};
  \node[tnode, line width=1.1pt] (t4) at (\xL,\yC) {$y$: negative};

  \foreach \nd in {t1,t2} \draw[inedge] (x.south -| \nd) -- (\nd);
  \draw[tedge] (t1) -- (t3);  \draw[tedge] (t2) -- (t3);
  \draw[tedge] (t3) -- (t4);

  \node[snode] (s1) at (\xR-\dx,\yA) {$c_1$: food\\ good};
  \node[snode] (s2) at (\xR+\dx,\yA) {$c_2$: service\\ slow};
  \node[snode] (s3) at (\xR,\yB)     {$c_3$: visit\\ soured};
  \node[snode, line width=1.1pt] (s4) at (\xR,\yC) {$\hat y$: negative};

  \foreach \nd in {s1,s2} \draw[inedge] (x.south -| \nd) -- (\nd);
  \draw[sedge] (s1) -- (s3);  \draw[sedge] (s2) -- (s3);
  \draw[sedge] (s3) -- (s4);

  \node[note] at (\xL,\yF) {$R_3$ is drawn\\ from $R_1$ and $R_2$};
  \node[note] at (\xR,\yF) {$f_3$ reads $x$ and its parents:\\
    $h_3 = f_3\big(\mathrm{concat}(h_0, h_1, h_2)\big)$};
\end{tikzpicture}
\vspace{-2mm}
\caption{\textbf{Mirroring a teacher's reasoning DAG into a Graph of Concept
Predictors, shown on a Yelp review.} Left: the LLM teacher judges the food
($R_1$) and the service ($R_2$), combines them into an overall impression
($R_3$), and labels the review negative. Right: GCP gives each conclusion its
own concept predictor $f_j$ and wires the predictors with the same edges;
every $f_j$ also reads $x$. The teacher's
intermediate answers serve as labels for the matching concepts $c_j$.}
\label{fig:case1}
\vspace{-3mm}
\end{figure}

To address these challenges, we propose Graph of Concept Predictors (GCP), a
reasoning-aware active distillation framework for small discriminative
models. As shown in Figure~\ref{fig:case1}, GCP explicitly represents the
teacher's reasoning process as a structured, directed acyclic graph and uses
it to guide both sample selection and model retraining. To address sample
selection, GCP scores each text by uncertainty, gradient diversity, and
coverage at every concept node, weighted by centrality, rather than by
final-output uncertainty alone, targeting the reasoning steps where the
student is likely to err. To address retraining, GCP decomposes the
student into modular sub-components aligned with reasoning stages and
performs targeted retraining on the modules most responsible for downstream
errors. This design enables dense, interpretable supervision of reasoning
while preserving the efficiency advantages of small discriminative models,
making reasoning-aware distillation practical at scale.

We summarize our main contributions as follows:
(a) We proposed \textbf{GCP}, an \textbf{active distillation} framework that
mirrors a teacher's reasoning graph with a learnable \textbf{Graph of Concept
Predictors}, enabling structured and interpretable reasoning transfer.
(b) We designed a \textbf{graph-aware acquisition strategy} that prioritizes
samples informative for learning concept dependencies and their downstream
effects.
(c) We introduced a \textbf{sub-module retraining} strategy that selectively
updates the most influential concept predictors while keeping other modules
fixed.
(d) We conducted \textbf{extensive experiments} on eight NLP classification
benchmarks, demonstrating that \textsc{GCP} achieves stronger performance
under limited annotation budgets with more stable and interpretable training
dynamics.

\vspace{-1mm}
\section{Related Work}
\subsection{Concept-Based Reasoning}


Concept Bottleneck Models (CBMs)~\cite{Koh2020ConceptBM} are the canonical
concept-based predictors, predicting each concept from the input alone and
the label from the concepts. Semi-supervised CBMs~\cite{Hu2024SemiSupervisedCB}
reduce concept annotation cost, while energy-based, relational, and graph
CBMs~\cite{Xu2024EnergyBasedCB,Barbiero2023RelationalCB,Xu2025GraphCB} model
dependencies among concepts. Reasoning distillation transfers free-text
rationales or question decompositions from a teacher to a
student~\cite{Hsieh2023DistillingSO,Shridhar2022DistillingRC}. 
These works
keep concepts as a \emph{flat} bottleneck, learn their dependencies
\emph{jointly} within one network, or transfer reasoning as
\emph{unstructured} text, rather than distilling a teacher-elicited concept
\emph{DAG} whose node-specific predictors localize student errors to
individual concepts for acquisition and retraining.

\vspace{-1.5mm}
\subsection{Large Language Models for Annotation}
LLMs annotate unlabeled text through zero- or few-shot
prompting~\cite{Brown2020LanguageMA,Kojima2022LargeLM}, and their labels can
train smaller task models at a fraction of the human labeling
cost~\cite{Wang2021WantTR}. LLM annotation has since been evaluated and
surveyed across NLP tasks and languages~\cite{Bhat2023LargeLM,Tan2024LargeLM}.
Chain-, tree-, and graph-of-thought
prompting~\cite{Wei2022ChainOT,Yao2023TreeOT,Yao2023BeyondCE} additionally
elicits the intermediate steps behind each answer.
These works
annotate each instance with a label alone, or elicit reasoning whose
steps and structure are regenerated per query, rather than annotating
every instance on a task-level concept DAG elicited once and held fixed,
whose node values supervise individual concept predictors and expose where
the student departs from the teacher.

\vspace{-1.5mm}
\subsection{Active Learning with LLM as Annotator}
Active learning selects examples by predictive
uncertainty~\cite{Lewis1995ASA,Kirsch2019BatchBALDEA}, representation
coverage~\cite{Sener2017ActiveLF,Citovsky2021BatchAL}, or gradient-embedding
diversity~\cite{Ash2019DeepBA}. Recent systems replace the human oracle with
an LLM annotator or decouple LLM-based selection from student
training~\cite{Xiao2023FreeALTH,Zhang2023LLMaAAML}. These criteria score a
sample by its final-label distribution or a single embedding, so they cannot
tell which reasoning step the sample would correct, and each round retrains
the whole student. GCP computes all three criteria per concept node, weighted
by graph centrality (Section~\ref{sec:acquisition}), and reuses the same graph
to retrain only the concept predictors most responsible for the loss
(Section~\ref{sec:submodule_retraining}).

\vspace{-1mm}
\section{Proposed Method}\vspace{-1mm}
\label{sec:method}
Given a classification task, an unlabeled pool, and an LLM teacher with an
annotation budget, GCP trains a small student that predicts labels through
the teacher's intermediate concepts. It consists of four components. First,
concept graph construction elicits the teacher's reasoning as a directed
acyclic graph (DAG) of concepts and keeps only the parts that the data
support. Next, a Graph of Concept Predictors mirrors this graph with one
predictor per concept. Graph-aware acquisition then uses the student's
concept-level predictions to choose texts for annotation, and finally
sub-module retraining updates only the predictors responsible for most of
the loss and freezes the rest. The central idea is that one concept graph indexes the
student's modules, the teacher's supervision, the acquisition scores, and the
retraining units. Figure~\ref{fig:framework} shows the loop.

\vspace{-1.5mm}
\subsection{Problem Formulation}
\label{sec:problem}
We study active distillation of an LLM teacher into a small text
classifier. Given a task definition, a label set, an unlabeled pool, and a
teacher that annotates each queried text with a label and concept values, the
goal is a student that labels unseen texts without the teacher. Two budgets
constrain the problem. The student has at most $M$ parameters and far below the teacher's size, so serving it is cheap. Its
training data costs at most $N$ teacher tokens. Performance is measured by
test accuracy.

Let $x$ denote a text, $y\in\mathcal{Y}$ its label, and $\mathcal{U}$ the
unlabeled pool. Concepts and their dependencies form a DAG $G=(V,E)$ with
$V=\{c_1,\dots,c_K,y\}$, where $\mathrm{pa}(j)$ denotes the parents of node
$j$, and $\mathcal{D}$ holds the annotated examples $(x,\mathbf{c},y)$ with
concept values $\mathbf{c}$. We seek
\begin{equation}
\begin{aligned}
\max\nolimits_{\mathcal{D}}\;\;&
\Pr\nolimits_{(x,y)\sim P}\big[\hat y(x;\theta_{\mathcal{D}})=y\big]\\
\text{s.t.}\;\;&
\mathrm{tok}(\mathcal{D})\le N,\quad |\theta_{\mathcal{D}}|\le M,
\end{aligned}
\label{eq:problem}
\end{equation}
where $\theta_{\mathcal{D}}$ is the student trained on $\mathcal{D}$ (texts
from $\mathcal{U}$) and $\hat y$ its prediction; $\mathrm{tok}(\mathcal{D})$
counts all teacher tokens, graph construction and retraining included, and
$|\theta_{\mathcal{D}}|$ counts all student parameters, any pretrained encoder
included. Since inference runs the student alone, $M$ also caps the
per-text inference cost. The probability is over the
task distribution $P$, estimated on a test set. We build
$\mathcal{D}$ in rounds of size $B$.

With the student's size fixed by $M$, accuracy must come from how the $N$
teacher tokens are spent and how the student learns from them. Solving
Eq.~\eqref{eq:problem} therefore requires \emph{concept-level supervision}, a
fixed concept set that the teacher annotates and the student mirrors with one
trainable unit per concept; \emph{reasoning-aware selection}, which scores a
text by where in the reasoning the student is uncertain; and \emph{localized
retraining}, which attributes the final loss to individual concept predictors
and updates only those. Existing methods meet these requirements only in part.
Concept bottleneck models predict concepts from the input alone or learn their
dependencies inside one
network~\cite{Koh2020ConceptBM,Barbiero2023RelationalCB,Xu2025GraphCB}, and
active-learning criteria score a text by its final-label distribution or a
single embedding and retrain the whole
student~\cite{Lewis1995ASA,Sener2017ActiveLF,Ash2019DeepBA}.

\vspace{-1.5mm}
\subsection{Overall Architecture}
\label{subsec:distill_goc_framework}

\begin{figure}[ht]
\vspace{-2mm}
\centerline{\includegraphics[width=\columnwidth]{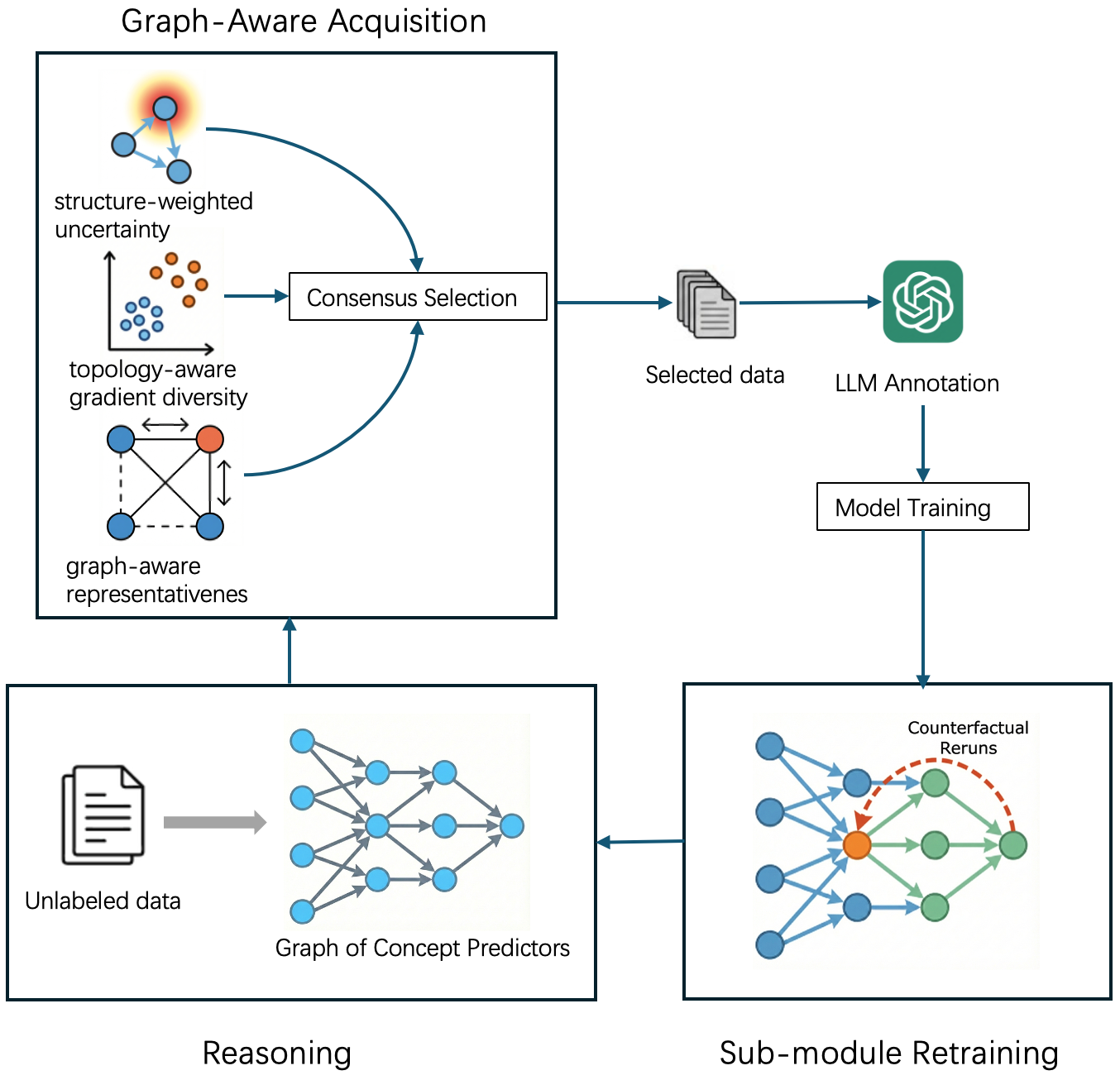}}
    \vspace{-2mm}
    \caption{
      \textbf{Overview of the GCP loop:} the student (lower left),
      graph-aware acquisition (top), and sub-module retraining by
      counterfactual reruns (lower right).
    }
    \label{fig:framework}
    \vspace{-2mm}
\end{figure}

These requirements determine the architecture of GCP. Concept graph
construction (Section~\ref{sec:graph_construction}) and the Graph of Concept
Predictors (Section~\ref{sec:gcp}) provide concept-level supervision,
graph-aware acquisition (Section~\ref{sec:acquisition}) provides
reasoning-aware selection, and sub-module retraining
(Section~\ref{sec:submodule}) provides localized retraining.

Reading Figure~\ref{fig:framework} clockwise from the lower-left panel,
the student runs on the unlabeled pool and outputs a distribution for every
concept and for the label. The upward arrow passes these node-level outputs to
acquisition, where three criteria nominate candidates and consensus selection
keeps the texts they agree on. The teacher annotates the selected texts with
labels and concept values, which join $\mathcal{D}$. In the
lower-right panel, a candidate predictor (orange) is corrected after its
parents, its descendants (green) are recomputed, and the dashed arrow returns
the resulting change in final loss to that predictor. The leftward arrow
closes the loop, and both lower panels show the same DAG, built once
beforehand.

This organization separates choosing texts to annotate from choosing
predictors to update and links the two through the concept graph. Because the teacher
annotates every concept, each query labels every predictor's output, and
retraining updates only the predictors causing most of the loss.

The architectural novelty lies in this shared interface. Existing active
distillation pipelines treat the student as one
network~\cite{Ash2019DeepBA,Zhang2023LLMaAAML}, whereas each concept in GCP
has its own supervision, acquisition terms, and retraining decision, so one
reasoning step can be repaired in isolation.
Sections~\ref{sec:graph_construction}--\ref{sec:execution} follow this flow.

\vspace{-1.5mm}
\subsection{Concept Graph Construction}
\label{sec:graph_construction}
From the task and teacher fixed in Section~\ref{sec:problem}, concept graph
construction builds, once per task, the graph that all later components
share. The teacher proposes concepts and dependencies, and its
annotations decide which survive. For the review in Figure~\ref{fig:case1},
an edge from the service judgment to the overall impression is kept only if
the two remain dependent given the text. Construction takes the task
definition, the label set, a few representative texts, and an initial
sample $\mathcal{X}_0\subset\mathcal{U}$, and returns a DAG that stays
fixed in all rounds. The annotations of $\mathcal{X}_0$ also form the initial
$\mathcal{D}$, so edges are selected on training data, never on test data.
No human writes concepts or edges.

\vspace{-1.5mm}
\paragraph{Stage 1: Concept Elicitation.}
The teacher lists, in order, the binary concepts a human annotator would
check, each citing the earlier concepts it depends on
(Appendix~\ref{sec:prompt_elicitation}). The list order defines a causal
ordering $\prec$, and only citations of earlier concepts become candidate
edges $E_{\mathrm{cand}}$.

\vspace{-1.5mm}
\paragraph{Stage 2: Concept Pruning.}
We remove near-duplicates by name similarity, keeping the earlier one in
$\prec$, have the teacher annotate the rest on $\mathcal{X}_0$,
and drop concepts with near-random annotations, $H(c)/\log 2>\eta$. The
survivors form $V$.

\vspace{-1.5mm}
\paragraph{Stage 3: Edge Induction.}
A candidate edge $c_i\to c_j$ between survivors is kept if
$\widehat{I}(c_i;c_j\mid x)/\min\{H(c_i),H(c_j)\}>\tau$ on
$\mathcal{X}_0$. All edges follow $\prec$, so $G$ is acyclic by
construction.

\looseness=-1
The entropy test removes concepts the teacher cannot annotate consistently,
and the edge test removes dependencies that add nothing given the text.

Compared with the teacher's proposed graph, construction rejects
a substantial share of the candidate edges and sizes the graph by task,
keeping fewer concepts on topic and sentiment benchmarks than on
multi-factor ones (Table~\ref{tab:graph_stats}). Its one-time token cost counts toward the
budget $N$ (Appendix~\ref{sec:concept_graph}).

\vspace{-1.5mm}
\subsection{Graph of Concept Predictors}
\label{sec:gcp}
\looseness=-1
Section~\ref{sec:graph_construction} produces $G$, and the Graph of
Concept Predictors turns it into a trainable student. Each predictor answers
one question about the text, reading the text and its parents' answers. In
Figure~\ref{fig:case1}, the student judges food and service, merges them into
an overall impression, and labels the review from it. Its per-node output
distributions feed acquisition and sub-module retraining. Unlike a concept
bottleneck model, it computes concepts along the teacher's dependencies with separate modules, so each reasoning step is
supervised and updated on its own.

A frozen encoder maps $x$ to $h_0=f_0(x)$, which every predictor reads.
Each node $j\in V$ has a predictor $f_j$ with parameters $\theta_j$,
evaluated in topological order:
\begin{equation}
h_j=f_j\big(\mathrm{concat}(h_0,\{h_i: i\in\mathrm{pa}(j)\})\big),
\label{eq:propagation}
\end{equation}
where $\mathrm{concat}$ joins $h_0$ and the parent outputs along the
feature dimension. Each $f_j$ is a small MLP with a softmax output
(Appendix~\ref{sec:experiment details}), so $h_j=p_j(x)$ is the distribution
of node $j$ given $x$ and its parents
(Section~\ref{sec:analysis}), and the label node yields $\hat y(x)$.
Annotated values replace $h_j$ by one-hot vectors $h_j^\star(x)$ of the
same size. Because each predictor has its own
parameters and output, the annotation of $c_j$ supervises $\theta_j$
directly, and an error at node $j$ is visible at its output rather than only
in $\hat y$.

Compared with a flat CBM and an end-to-end MLP, the DAG factorization lowers
the joint concept--label risk: for DAG-factorized data, the flat CBM's
excess risk over GCP is exactly
$\mathbb{E}[\mathrm{TC}(\mathbf{c}\mid x)]$, the concepts' conditional total
correlation, and the MLP's excess over the flat CBM is
$\mathbb{E}[I(y;\mathbf{c}\mid x)]$ (Theorem~\ref{thm:gcp_risk}), and
GCP outperforms both on every benchmark
(Table~\ref{tab:ablation-architecture}).

\vspace{-1.5mm}
\subsection{Graph-Aware Acquisition}
\label{sec:acquisition}
Graph-aware acquisition uses the node-level distributions from
Section~\ref{sec:gcp} to choose the $B$ texts annotated in each round. It
prefers texts on which the student is uncertain at central concepts, that would move the predictors in uncovered directions,
and that lie far from the annotated texts in concept space. In
Figure~\ref{fig:case1}, uncertainty about the overall impression $c_3$ counts
more than uncertainty about the food judgment $c_1$, because $c_3$ connects to
more nodes. Each criterion is computed per node and weighted by degree
centrality
$w_i=\deg(i)/\sum_{j\in V}\deg(j)$, so a text is scored by where in the
reasoning it is informative.

\vspace{-1.5mm}
\paragraph{Structure-Weighted Uncertainty.}
\label{sec:topo_entropy}
With $H_i(x)$ the entropy of $p_i(x)$, the structure-weighted uncertainty
(SWU) score is
\[
E_{\mathrm{unc}}(x)=\Big(\textstyle\sum_{i\in V} w_i H_i(x)^p\Big)^{1/p}\!,\quad p\!\ge\!1,
\]
and $\mathcal{S}^{\mathrm{SWU}}_{k}$ holds the $k$ highest-scoring texts in
$\mathcal{U}$. Raising $p$ moves $E_{\mathrm{unc}}$ from the weighted mean of
node entropies toward the largest one, so uncertainty at one central concept
is not averaged away.

\vspace{-1.5mm}
\paragraph{Topology-Aware Gradient Diversity.}
\label{sec:topo_grad}
For unannotated $x$, $z_i(x)$ is the gradient of node $i$'s loss with
respect to $\theta_i$ at the pseudo-label $\arg\max p_i(x)$, with parents
from Eq.~\eqref{eq:propagation}, as in BADGE~\cite{Ash2019DeepBA}. We
cluster $\mathcal{U}$ into $k$ groups by
\[
D_{\mathrm{grad}}(x,x')
=\Big(\textstyle\sum_{i\in V} w_i \|z_i(x)-z_i(x')\|_2^p\Big)^{1/p}
\]
and take the cluster medoids as $\mathcal{S}^{\mathrm{grad}}_k$. Texts are
close only if they would move the same predictors in similar directions, so
the medoids spread the batch over distinct updates.

\vspace{-1.5mm}
\paragraph{Graph-Aware Representativeness.}
\label{sec:graph_repr}
We compare node distributions by a weighted, symmetrized KL divergence,
\[
\begin{aligned}
D_{\mathrm{SKL}}(x,x')
&=\tfrac{1}{2}\textstyle\sum_{i\in V}w_i[d_i(x,x')+d_i(x',x)],\\[-1mm]
d_i(x,x')&=\mathrm{KL}\big(p_i(x)\,\|\,p_i(x')\big),
\end{aligned}
\]
and approximate by clustering the core set
\[
\mathcal{S}^{\mathrm{cover}}_k
=\arg\min_{\substack{S\subseteq\mathcal{U}\\|S|=k}}
\max_{x\in\mathcal{U}}\min_{x'\in\mathcal{D}\cup S}D_{\mathrm{SKL}}(x,x'),
\]
which bounds how far any unlabeled text lies from the annotated set in
concept-prediction space.

\vspace{-1.5mm}
\paragraph{Consensus Selection.}
\label{sec:consensus_sampling}
We select
$\mathcal{S}^\ast=\mathcal{S}^{\mathrm{SWU}}_k\cap\mathcal{S}^{\mathrm{grad}}_k\cap\mathcal{S}^{\mathrm{cover}}_k$,
the texts that are uncertain, diverse, and representative at once, and fill
any remaining slots up to $B$ from the union of the three sets by a
tie-breaking score such as $E_{\mathrm{unc}}(x)$.

Compared with Entropy Sampling~\cite{Settles2009ActiveLearning},
BADGE~\cite{Ash2019DeepBA}, and CoreSet~\cite{Sener2017ActiveLF}, which apply
these ideas to the final-label distribution or a single embedding,
graph-aware acquisition applies them per node. Removing any criterion lowers
accuracy on every benchmark, and replacing the intersection with the
union costs more than any single removal (Table~\ref{tab:ablation-components}).

\begin{figure}[ht]
\centering
\begin{tikzpicture}[
    x=1cm, y=1cm,
    font=\scriptsize,
    box/.style={rounded corners=2.5pt, align=flush center, line width=0.6pt,
                text width=1.55cm, minimum height=0.62cm, inner sep=1.5pt},
    tnode/.style={box, draw=gcpTDraw, fill=gcpTFill},
    snode/.style={box, draw=gcpSDraw, fill=gcpSFill},
    enode/.style={box, draw=gcpEDraw, fill=gcpEFill, line width=0.9pt},
    hit/.style={dash pattern=on 2pt off 1.2pt},
    inedge/.style={-{Stealth[length=3.5pt,width=3pt]}, line width=0.6pt, draw=gray!70},
    tedge/.style={-{Stealth[length=3.5pt,width=3pt]}, line width=0.7pt, draw=gcpTDraw},
    sedge/.style={-{Stealth[length=3.5pt,width=3pt]}, line width=0.7pt, draw=gcpSDraw},
    eedge/.style={-{Stealth[length=3.5pt,width=3pt]}, line width=0.8pt, draw=gcpEDraw},
    note/.style={fill=gcpGray, rounded corners=2pt, inner sep=2.5pt, align=center},
  ]
  \def\xL{1.85}  \def\xR{5.85}  \def\dx{0.92}
  \def\yT{0}     \def\yX{-0.62}
  \def\yA{-1.5}  \def\yB{-2.45} \def\yC{-3.4} \def\yF{-4.25}

  \node[font=\footnotesize\bfseries, text=gcpTDraw] at (\xL,\yT) {Teacher};
  \node[font=\footnotesize\bfseries, text=gcpSDraw] at (\xR,\yT) {Student (GCP)};
  \node[single arrow, draw=gray!60, fill=gcpGray, minimum height=1.1cm,
        single arrow head extend=2pt, inner sep=1.2pt]
    (mirror) at (3.85,\yT) {mirror};
  \draw[line width=0.8pt, gray!70] (mirror.south) -- (3.85,\yF-0.4);

  \node[draw=gray!60, fill=white, rounded corners=2.5pt, inner sep=2.5pt,
        line width=0.5pt, minimum width=7.5cm] (x) at (3.85,\yX)
    {\textbf{Input $x$}:\ \emph{``Great food, but we waited an hour.''}};

  \node[tnode] (t1) at (\xL-\dx,\yA) {$R_1$: food\\ good};
  \node[tnode] (t2) at (\xL+\dx,\yA) {$R_2$: service\\ slow};
  \node[tnode] (t3) at (\xL,\yB)     {$R_3$: visit\\ soured};
  \node[tnode, line width=1.1pt] (t4) at (\xL,\yC) {$y$: negative};

  \foreach \nd in {t1,t2} \draw[inedge] (x.south -| \nd) -- (\nd);
  \draw[tedge] (t1) -- (t3);  \draw[tedge] (t2) -- (t3);
  \draw[tedge] (t3) -- (t4);

  \node[snode] (s1) at (\xR-\dx,\yA) {$c_1$: food\\ good};
  \node[enode] (s2) at (\xR+\dx,\yA) {$c_2$: service\\ fine\ \textcolor{gcpEDraw}{$\times$}};
  \node[snode, hit] (s3) at (\xR,\yB)     {$c_3$: visit\\ soured};
  \node[snode, hit, line width=1.1pt] (s4) at (\xR,\yC) {$\hat y$: negative};

  \foreach \nd in {s1,s2} \draw[inedge] (x.south -| \nd) -- (\nd);
  \draw[sedge] (s1) -- (s3);  \draw[eedge] (s2) -- (s3);
  \draw[eedge] (s3) -- (s4);

  \node[note] at (\xL,\yF) {$P(\text{negative}) = 0.93$\\ loss $= 0.07$};
  \node[note] at (\xR,\yF) {$P(\text{negative}) = 0.56$\\ loss $= 0.58$};
\end{tikzpicture}
    \vspace{-2mm}
    \caption{\textbf{A correct but under-confident prediction, on the review
    from Figure~\ref{fig:case1}.} The student's service predictor $f_2$
    misses the complaint about waiting (red), and the error passes to $c_3$
    and $\hat y$ (dashed). The student still predicts \emph{negative}, but with
    lower confidence and higher loss.}
    \label{fig:case2}
    \vspace{-2mm}
\end{figure}
\vspace{-1.5mm}
\subsection{Sub-module Retraining}
\label{sec:submodule}
\label{sec:submodule_retraining}
Once a round's annotations join $\mathcal{D}$, sub-module retraining updates
only the $b$ predictors that cause most of the final loss and freezes the
rest. It asks how much the
final loss drops when one predictor is corrected after its parents. In
Figure~\ref{fig:case2}, $f_2$
misses the complaint about waiting, and the error passes through $c_3$ and
raises the final loss. Correcting $c_2$ lets $c_3$ and $\hat y$ be
recomputed from the right judgment, which marks $f_2$ for retraining. Its key distinction is attributing loss through interventions on
the concept graph, separating a predictor's own error from inherited ones.

For an intervention set $S\subseteq V$, the nodes in $S$ take their annotated
values $h_j^\star(x)$ and the others are recomputed in topological order:
\[
\tilde h_j^S(x)=
\begin{cases}
h_j^\star(x), & j\in S,\\[2pt]
f_j\big(\mathrm{concat}(h_0,\tilde h^S_{\mathrm{pa}(j)}(x))\big), & j\notin S.
\end{cases}
\]
Let $\mathcal{L}^{\mathrm{do}(S)}(x)$ be the loss of the resulting label
prediction. The \emph{parent-corrected} rerun gives
$\mathcal{L}^{\mathrm{pa}}_i=\mathcal{L}^{\mathrm{do}(\mathrm{pa}(i))}$ and
the \emph{parent+node-corrected} rerun gives
$\mathcal{L}^{\mathrm{pa}+i}_i=\mathcal{L}^{\mathrm{do}(\mathrm{pa}(i)\cup\{i\})}$,
which define the influence score
\[
\Delta_i=\tfrac{1}{|\mathcal{D}|}\textstyle\sum_{(x,\mathbf{c},y)\in\mathcal{D}}
\big[\mathcal{L}^{\mathrm{pa}}_i(x)-\mathcal{L}^{\mathrm{pa}+i}_i(x)\big].
\]
The retrained set $\mathcal{M}^\star$ is the set $\mathcal{M}\subseteq V$ of
$b$ nodes that maximizes $\sum_{i\in\mathcal{M}}\Delta_i$. Each $f_i$,
$i\in\mathcal{M}^\star$, continues from its current weights and minimizes
$F_i$ (Eq.~\eqref{eq:training}) on $\mathcal{D}$, and the other predictors
stay frozen. For a selected concept node, the teacher also generates short
texts labeled with $c_i$ and its parents
(Appendix~\ref{sec:prompt_regeneration}), added to the training data of
$f_i$ (Algorithm~\ref{alg:submodule_retrain}). Both reruns start from corrected parents, so $\Delta_i$ measures the loss
that node $i$ adds by itself, and the generated texts, which carry no task
label, never train the label predictor.

\looseness=-1
Unlike whole-student retraining, which ignores where the loss originates,
sub-module retraining focuses each round's updates on the $b$ predictors
with the largest attributed loss. Taking
the top $b$ scores solves the additive selection problem exactly
(Theorem~\ref{thm:optimal}), and removing this step causes the largest
ablation drop on every benchmark (Table~\ref{tab:ablation-components}).

\vspace{-1.5mm}
\subsection{Training and End-to-End Execution}
\label{sec:execution}
The student is trained with teacher forcing: each predictor receives the
annotated values of its parents, so the objective separates across nodes,
\begin{align}
F(\theta)&=\textstyle\sum_{j\in V}\lambda_j F_j(\theta_j),
\label{eq:training}\\
F_j(\theta_j)&=\textstyle\frac{1}{|\mathcal{D}|}\sum_{\mathcal{D}}
-\log p_j\big(v_j\!\mid\! h_0,h^\star_{\mathrm{pa}(j)}(x);\theta_j\big),\nonumber
\end{align}
where $v_j$ is the annotated value of node $j$ and $\lambda_j>0$ its weight.
We minimize $F$ with AdamW (Appendix~\ref{sec:experiment details}) over
all predictors once, on the annotated $\mathcal{X}_0$; later rounds
minimize $F_i$ only for $i\in\mathcal{M}^\star$, leaving the frozen
predictors' losses unchanged since $F$ separates across nodes.
Algorithm~\ref{alg:gcp} (Appendix~\ref{sec:alg_gcp}) gives the complete
loop, alternating acquisition, annotation, and sub-module retraining until
the budget $N$ is spent. At inference, the student evaluates Eq.~\eqref{eq:propagation} alone.

\vspace{-1.5mm}
\subsection{Theoretical Analysis}
\label{sec:analysis}
Here we analyze the training objectives through which GCP approximates
Eq.~\eqref{eq:problem}: the DAG factorization lowers the joint
concept--label risk, the teacher-forced objective converges linearly, and
sub-module retraining selects its predictors optimally in polynomial time,
with one concept graph underlying all three. Due to limited space, full
statements and proofs are deferred to Appendix~\ref{sec:proof}.

\vspace{-1mm}
\begin{theorem}[Joint-Risk Advantage]
\label{thm:gcp_risk}
Let $p^\star(y,\mathbf{c}\mid x)$ factorize over $G$, and let
$\mathcal{R}_{\mathrm{M}}$ be the lowest expected joint loss
$-\log p(y,\mathbf{c}\mid x)$ attainable by GCP, a flat CBM
($p(y\mid x,\mathbf{c})\prod_jp(c_j\mid x)$), or an MLP
($p(y\mid x)\prod_jp(c_j\mid x)$). Then
$\mathcal{R}_{\mathrm{CBM}}-\mathcal{R}_{\mathrm{GCP}}
=\mathbb{E}[\mathrm{TC}(\mathbf{c}\mid x)]$ and
$\mathcal{R}_{\mathrm{MLP}}-\mathcal{R}_{\mathrm{CBM}}
=\mathbb{E}[I(y;\mathbf{c}\mid x)]$, where $\mathrm{TC}$ is the concepts'
conditional total correlation.
\end{theorem}

\vspace{-1mm}
\begin{theorem}[Linear Convergence]
\label{thm:gcp_conv}
Under Assumption~\ref{asm:teacher_forcing}, let each $F_j$ in
Eq.~\eqref{eq:training} be $L_j$-smooth and satisfy a
Polyak--\L{}ojasiewicz (PL) inequality~\citep{karimi2016linear} with
constant $\mu_j$ near a minimizer $\theta^\star$. Gradient descent with step
$1/L$, $L=\max_j\lambda_jL_j$, satisfies
$F(\theta_t)-F(\theta^\star)\le(1-\mu_{\mathrm{GCP}}/L)^t
(F(\theta_0)-F(\theta^\star))$, where
$\mu_{\mathrm{GCP}}=\min_j\lambda_j\mu_j$.
\end{theorem}

\vspace{-1mm}
\begin{proposition}[Ordering of Convergence Constants]
\label{prop:mu_order}
If, in addition, conditioning on parents does not lower any node's PL
constant (Assumption~\ref{asm:curvature}) and the weights $\lambda_j$ are
shared, then $\mu_{\mathrm{GCP}}\ge\mu_{\mathrm{CBM}}\ge\mu_{\mathrm{MLP}}$.
\end{proposition}

\vspace{-1mm}
\begin{theorem}[Optimal Retraining Selection]
\label{thm:optimal}
\label{thm:timecomplexity}
The $b$ nodes with the largest $\Delta_i$ maximize
$\sum_{i\in\mathcal{M}}\Delta_i$ over all $\mathcal{M}\subseteq V$ with
$|\mathcal{M}|=b$, and computing every $\Delta_i$ takes
$\mathcal{O}(|\mathcal{D}||V|^2)$ time per round in the worst case.
\end{theorem}

\vspace{-1mm}
\begin{remark*}[Scope]
These guarantees concern training objectives, whereas the accuracy gains
reported next are empirical. Estimated PL constants support
Assumption~\ref{asm:curvature} (Table~\ref{tab:pl_constants}), and per-round
runtime stays within a small constant factor of BADGE's
(Table~\ref{tab:runtime}).
\end{remark*}

\vspace{-1mm}
\section{Experiments}
\begin{table*}[b!]
\vspace{-2mm}
\centering
\small
\setlength{\tabcolsep}{4pt}
\renewcommand{\arraystretch}{1.0}
\caption{\textbf{Baseline comparison under a fixed 20\% token budget.}
All methods are evaluated in a pool-based active learning setting and spend
the same number of teacher tokens, so the label-only baselines annotate more
texts than GCP. Best results are in \textbf{bold}.}
\label{tab:baseline_20}
\vspace{-1.5mm}
\begin{tabular}{lcccccccc}
\toprule
\textbf{Method} &
\textbf{MNLI} &
\textbf{GoEmotions} &
\textbf{SemEval} &
\textbf{MIMIC-III} &
\textbf{AG News} &
\textbf{Amazon} &
\textbf{IMDB} &
\textbf{Yelp} \\
\midrule

Random
& 51.21 & 23.06 & 30.33 & 75.21 & 82.87 & 86.60 & 79.21 & 89.19 \\

\midrule
\multicolumn{9}{l}{\textit{Classic uncertainty- and diversity-based baselines}} \\

Least Confidence
& 57.42 & 25.01 & 35.18 & 80.32 & 83.05 & 88.94 & 82.37 & 91.84 \\

Entropy Sampling
& 59.30 & 25.98 & 37.46 & 81.51 & 83.27 & 89.63 & 83.41 & 92.56 \\

Disagreement (Vote Entropy)
& 58.91 & 25.44 & 36.87 & 81.03 & 83.11 & 89.21 & 83.02 & 92.19 \\

\midrule
\multicolumn{9}{l}{\textit{Representation- and gradient-based strong baselines}} \\

CoreSet (k-center)
& 59.60 & 25.80 & 36.42 & 82.82 & 83.23 & 89.32 & 83.37 & 92.03 \\

ALPS
& 60.12 & 26.21 & 38.94 & 83.41 & 83.56 & 89.87 & 84.26 & 92.48 \\

CAL
& 60.83 & 26.54 & 40.58 & 84.71 & 83.91 & 90.54 & 87.62 & 93.12 \\

BADGE
& 60.91 & 26.01 & 39.53 & 83.78 & 83.31 & 89.81 & 85.91 & 92.62  \\

\midrule
\textbf{GCP (Ours)}
& \textbf{62.47} & \textbf{27.89} & \textbf{42.96} & \textbf{86.32}
& \textbf{84.73} & \textbf{91.62} & \textbf{88.94} & \textbf{94.21} \\
\bottomrule
\end{tabular}
\vspace{-2mm}
\end{table*}

\begin{figure*}[t!]
\small
    \centering
    \includegraphics[width=\textwidth]{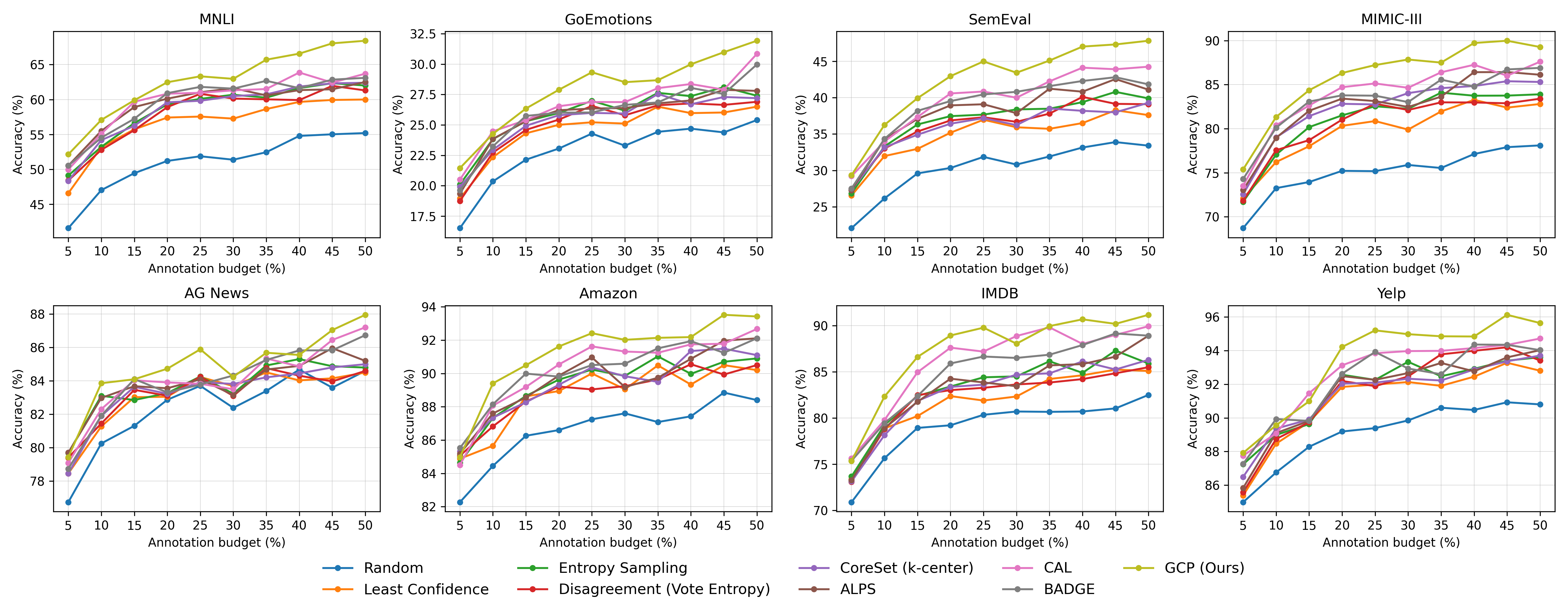}
    \vspace{-5.5mm}
    \caption{
     \textbf{Performance curves of different sample selection methods for
     active learning.} The y-axis denotes classification accuracy, and the
     x-axis denotes the teacher-token budget as a percentage of the tokens
     needed to label the full training set with labels only. At the same
     budget, GCP annotates fewer texts than the baselines because each of its
     queries also returns concept values.
    }
    \label{fig:draw_trade_off}
    \vspace{-1mm}
\end{figure*}
\begin{table*}[t!]
\centering
\small
\setlength{\tabcolsep}{6pt}
\caption{\textbf{Architecture ablation.} Comparison of GCP with a flat CBM
(no dependencies) and an end-to-end MLP at 50\% annotation budget.}
\vspace{-2mm}
\begin{tabular}{lcccccccc}
\toprule
\textbf{Architecture} & \textbf{MNLI} & \textbf{GoEmotions} & \textbf{SemEval} & \textbf{MIMIC-III} & \textbf{AG News} & \textbf{Amazon} & \textbf{IMDB} & \textbf{Yelp} \\
\midrule
\textbf{GCP}
& \textbf{68.41} & \textbf{31.92} & \textbf{47.83} & \textbf{89.27}
& \textbf{87.96} & \textbf{93.42} & \textbf{91.18} & \textbf{95.63} \\

Flat CBM
& 65.73 & 30.84 & 44.26 & 86.12
& 87.21 & 92.67 & 89.94 & 94.71 \\

End-to-End MLP
& 63.18 & 29.97 & 41.83 & 84.03
& 86.74 & 92.11 & 88.92 & 94.02 \\

\bottomrule
\end{tabular}
\label{tab:ablation-architecture}
\vspace{-3.5mm}
\end{table*}
\begin{table*}[t!]
\centering
\small
\setlength{\tabcolsep}{3pt}
\caption{\textbf{Ablation study on major components of \textsc{GCP}.}
Accuracy (\%) at 50\% annotation budget across eight datasets. Each row
disables one component; the full model includes all components.}
\vspace{-2mm}
\begin{tabular}{lcccccccc}
\toprule
\textbf{Setting} & \textbf{MNLI} & \textbf{GoEmotions} & \textbf{SemEval} & \textbf{MIMIC-III} & \textbf{AG News} & \textbf{Amazon} & \textbf{IMDB} & \textbf{Yelp} \\
\midrule
\textsc{GCP} (Full)
& \textbf{68.41} & \textbf{31.92} & \textbf{47.83} & \textbf{89.27}
& \textbf{87.96} & \textbf{93.42} & \textbf{91.18} & \textbf{95.63} \\

w/o SWU
& 66.82 & 30.74 & 45.61 & 87.43
& 87.21 & 92.81 & 90.32 & 94.91 \\

w/o Grad
& 66.41 & 30.28 & 45.02 & 87.11
& 87.08 & 92.64 & 89.97 & 94.74 \\

w/o Coverage
& 67.02 & 30.91 & 46.13 & 87.86
& 87.35 & 92.93 & 90.51 & 95.02 \\

w/o Intersection (Union + tie-break)
& 65.94 & 30.12 & 44.58 & 86.73
& 86.89 & 92.41 & 89.63 & 94.38 \\

w/o Sub-module Retraining
& 64.87 & 29.96 & 43.21 & 85.02
& 86.71 & 92.18 & 89.11 & 94.09 \\

\bottomrule
\end{tabular}
\label{tab:ablation-components}
\vspace{-5mm}
\end{table*}
\vspace{-1.5mm}
\subsection{Results and Analysis}
This section evaluates GCP against eight active-learning acquisition
strategies on eight real-world benchmarks. All methods spend the same teacher
tokens; a $p\%$ budget is $p\%$ of the tokens that label-only annotation of
the full training set needs, so the baselines annotate more texts than GCP.
Performance trends from a 5\% to
50\% budget appear in Figure~\ref{fig:draw_trade_off}, while
Table~\ref{tab:baseline_20} reports results at a fixed 20\% budget
(details in Appendix~\ref{sec:experiment details}).

\vspace{-1.5mm}
\paragraph{Overall Performance under Fixed Annotation Budget.}
As shown in Table~\ref{tab:baseline_20}, GCP delivers the best performance
across all eight datasets under a fixed 20\% annotation budget, outperforming
both uncertainty-based methods and strong gradient- and
representation-based baselines. On MNLI, GCP reaches 62.47\% accuracy,
beating the strongest baseline (BADGE at 60.91\%) by 1.56 points; on SemEval
it improves by 2.38 points (42.96\% vs.\ 40.58\%). On the clinically complex
MIMIC-III task, GCP achieves 86.32\%, exceeding CAL (84.71\%) by 1.61
points. These gains are consistent with graph-aware acquisition selecting
samples by uncertainty at central concept nodes rather than by output
uncertainty alone.
\vspace{-1.5mm}
\paragraph{Annotation Efficiency and Learning Dynamics.}
Figure~\ref{fig:compute_cost_vs_annotation_scale}
(Appendix~\ref{sec:runtime}) shows that GCP scales far more efficiently
than the LLaMA-3-70B teacher. As the number of labeled samples grows from $10^4$ to $10^6$,
the teacher's total compute cost rises from approximately $10^{18}$ to
$10^{20}$ FLOPs, whereas GCP remains in the range of
$10^{12}$--$10^{14}$ FLOPs. This corresponds to a
$\sim10^{5}$--$10^{6}\times$ reduction at every scale, enabling
iterative training without repeated full-scale teacher inference.

Figure~\ref{fig:draw_trade_off} further shows that GCP improves faster
throughout the 5\%--50\% budget range, leading competing methods at early,
middle, and late active-learning stages. The margin is largest at low
budgets, where output-uncertainty methods can be unstable and
gradient-based methods can over-focus on local decision boundaries. The
two GCP mechanisms address complementary parts of this problem:
graph-aware acquisition selects examples that are globally informative
with respect to the concept DAG, while counterfactual-guided retraining
updates only the concept predictors with the greatest downstream influence.
Freezing the remaining predictors stabilizes optimization and converts
sparse annotations into sustained gains across rounds.
\vspace{-1.5mm}
\subsection{Ablation Study}
We ablate each component of GCP at a 50\% annotation budget
(Table~\ref{tab:ablation-components}). Removing either major acquisition
signal reduces accuracy: MNLI drops from 68.41 to 66.82 without SWU and
66.41 without gradient diversity, while SemEval drops from 47.83 to 45.61
and 45.02. This confirms that structure-weighted uncertainty and
topology-aware gradient diversity contribute complementary information. The
pipeline components show the same pattern. On MNLI, 68.41 falls to 65.94 without
intersection and 64.87 without sub-module retraining; on MIMIC-III, 89.27
falls to 86.73 and 85.02; and on Yelp, 95.63 falls to 94.38 and 94.09.

The corresponding token--accuracy analysis of targeted retraining is
reported in Appendix~\ref{sec:token_accuracy}.
\vspace{-1.5mm}
\section{Conclusion}\vspace{-1.5mm}
This paper introduced GCP, an active distillation framework that transfers
LLM reasoning into an explicit Graph of Concept Predictors for interpretable,
concept-level supervision and diagnosis. GCP pairs graph-aware acquisition
with counterfactual sub-module retraining, which updates only the predictors
that most influence downstream loss and freezes the rest. Across eight
benchmarks, GCP outperforms strong active-learning baselines under limited
annotation budgets, and ablations show that acquisition and retraining are
complementary.

\section*{Limitations}
Our evaluation covers eight English text classification benchmarks. We have
not applied GCP to generation, structured prediction, or non-English and
multilingual tasks, where the concepts a teacher proposes and the
dependencies among them may take different forms.

All experiments use LLaMA-3-70B as the teacher. GCP interacts with the
teacher only through prompts (Appendix~\ref{sec:prompts}), so another
open-weight or proprietary model can serve as the teacher without changes
to the method, but we did not compare teachers of different families or
sizes.

Every concept node is binary. Properties that are naturally graded, such as
emotion intensity or the severity of a clinical finding, are therefore
expressed as binary concepts. Categorical or continuous nodes would require
different predictor heads and different pruning and edge criteria.

The concept graph is a DAG built once per task and kept fixed across rounds.
This matches the pool-based setting we study, in which the unlabeled pool is
available from the start. Streaming data with distribution shift may call
for updating the graph between rounds, and mutually reinforcing factors
would need cyclic or undirected structures, which GCP does not currently
support.

Finally, the MIMIC-III results measure accuracy on a retrospective clinical
benchmark. The distilled student has not been validated for clinical use,
and deployment in clinical decision support would require expert review of
both the induced graph and individual predictions.

\bibliography{custom}

@article{Ash2019DeepBA,
  title={Deep Batch Active Learning by Diverse, Uncertain Gradient Lower Bounds},
  author={Jordan T. Ash and Chicheng Zhang and Akshay Krishnamurthy and John Langford and Alekh Agarwal},
  journal={ArXiv},
  year={2019},
  volume={abs/1906.03671},
  url={https://api.semanticscholar.org/CorpusID:182953134}
}

@article{Koh2020ConceptBM,
  title={Concept Bottleneck Models},
  author={Pang Wei Koh and Thao Nguyen and Yew Siang Tang and Stephen O. Mussmann and Emma Pierson and Been Kim and Percy Liang},
  journal={ArXiv},
  year={2020},
  volume={abs/2007.04612},
  url={https://api.semanticscholar.org/CorpusID:220424448}
}

@inproceedings{Zhang2015CharacterlevelCN,
  title={Character-level Convolutional Networks for Text Classification},
  author={Xiang Zhang and Junbo Jake Zhao and Yann LeCun},
  booktitle={Neural Information Processing Systems},
  year={2015},
  url={https://api.semanticscholar.org/CorpusID:368182}
}

@inproceedings{Maas2011LearningWV,
  title={Learning Word Vectors for Sentiment Analysis},
  author={Andrew L. Maas and Raymond E. Daly and Peter T. Pham and Dan Huang and A. Ng and Christopher Potts},
  booktitle={Annual Meeting of the Association for Computational Linguistics},
  year={2011},
  url={https://api.semanticscholar.org/CorpusID:1428702}
}

@article{Brown2020LanguageMA,
  title={Language Models are Few-Shot Learners},
  author={Tom B. Brown and Benjamin Mann and Nick Ryder and Melanie Subbiah and Jared Kaplan and Prafulla Dhariwal and Arvind Neelakantan and Pranav Shyam and Girish Sastry and Amanda Askell and Sandhini Agarwal and Ariel Herbert-Voss and Gretchen Krueger and Thomas Henighan and Rewon Child and Aditya Ramesh and Daniel M. Ziegler and Jeff Wu and Clemens Winter and Christopher Hesse and Mark Chen and Eric Sigler and Ma-teusz Litwin and Scott Gray and Benjamin Chess and Jack Clark and Christopher Berner and Sam McCandlish and Alec Radford and Ilya Sutskever and Dario Amodei},
  journal={ArXiv},
  year={2020},
  volume={abs/2005.14165},
  url={https://api.semanticscholar.org/CorpusID:218971783}
}

@article{Lewis1995ASA,
  title={A sequential algorithm for training text classifiers: corrigendum and additional data},
  author={David D. Lewis},
  journal={SIGIR Forum},
  year={1995},
  volume={29},
  pages={13-19},
  url={https://api.semanticscholar.org/CorpusID:15704538}
}

@inproceedings{Citovsky2021BatchAL,
  title={Batch Active Learning at Scale},
  author={Gui Citovsky and Giulia DeSalvo and Claudio Gentile and Lazaros Karydas and Anand Rajagopalan and Afshin Rostamizadeh and Sanjiv Kumar},
  booktitle={Neural Information Processing Systems},
  year={2021},
  url={https://api.semanticscholar.org/CorpusID:236635499}
}

@article{Kirsch2019BatchBALDEA,
  title={BatchBALD: Efficient and Diverse Batch Acquisition for Deep Bayesian Active Learning},
  author={Andreas Kirsch and Joost R. van Amersfoort and Yarin Gal},
  journal={ArXiv},
  year={2019},
  volume={abs/1906.08158},
  url={https://api.semanticscholar.org/CorpusID:195069328}
}

@article{Xiao2023FreeALTH,
  title={FreeAL: Towards Human-Free Active Learning in the Era of Large Language Models},
  author={Rui Xiao and Yiwen Dong and Junbo Zhao and Runze Wu and Minmin Lin and Gang Chen and Haobo Wang},
  journal={ArXiv},
  year={2023},
  volume={abs/2311.15614},
  url={https://api.semanticscholar.org/CorpusID:265456919}
}

@inproceedings{Zhang2023LLMaAAML,
  title={LLMaAA: Making Large Language Models as Active Annotators},
  author={Ruoyu Zhang and Yanzeng Li and Yongliang Ma and Ming Zhou and Lei Zou},
  booktitle={Conference on Empirical Methods in Natural Language Processing},
  year={2023},
  url={https://api.semanticscholar.org/CorpusID:264814421}
}

@article{Hinton2015DistillingTK,
  title={Distilling the Knowledge in a Neural Network},
  author={Geoffrey E. Hinton and Oriol Vinyals and Jeffrey Dean},
  journal={ArXiv},
  year={2015},
  volume={abs/1503.02531},
  url={https://api.semanticscholar.org/CorpusID:7200347}
}

@article{Wang2021WantTR,
  title={Want To Reduce Labeling Cost? GPT-3 Can Help},
  author={Shuohang Wang and Yang Liu and Yichong Xu and Chenguang Zhu and Michael Zeng},
  journal={ArXiv},
  year={2021},
  volume={abs/2108.13487},
  url={https://api.semanticscholar.org/CorpusID:237363383}
}

@inproceedings{Tan2024LargeLM,
  title={Large Language Models for Data Annotation and Synthesis: A Survey},
  author={Zhen Tan and Dawei Li and Song Wang and Alimohammad Beigi and Bohan Jiang and Amrita Bhattacharjee and Mansooreh Karami and Jundong Li and Lu Cheng and Huan Liu},
  booktitle={Conference on Empirical Methods in Natural Language Processing},
  year={2024},
  url={https://api.semanticscholar.org/CorpusID:273901181}
}

@article{Siddhant2018DeepBA,
  title={Deep Bayesian Active Learning for Natural Language Processing: Results of a Large-Scale Empirical Study},
  author={Aditya Siddhant and Zachary Chase Lipton},
  journal={ArXiv},
  year={2018},
  volume={abs/1808.05697},
  url={https://api.semanticscholar.org/CorpusID:52039644}
}

@article{Bai2024BeyondEA,
  title={Beyond Efficiency: A Systematic Survey of Resource-Efficient Large Language Models},
  author={Guangji Bai and Zheng Chai and Chen Ling and Shiyu Wang and Jiaying Lu and Nan Zhang and Tingwei Shi and Ziyang Yu and Mengdan Zhu and Yifei Zhang and Carl Yang and Yue Cheng and Liang Zhao},
  journal={ArXiv},
  year={2024},
  volume={abs/2401.00625},
  url={https://api.semanticscholar.org/CorpusID:266693677}
}

@inproceedings{Williams2017ABC,
  title={A Broad-Coverage Challenge Corpus for Sentence Understanding through Inference},
  author={Adina Williams and Nikita Nangia and Samuel R. Bowman},
  booktitle={North American Chapter of the Association for Computational Linguistics},
  year={2017},
  url={https://api.semanticscholar.org/CorpusID:3432876}
}

@inproceedings{Demszky2020GoEmotionsAD,
  title={GoEmotions: A Dataset of Fine-Grained Emotions},
  author={Dorottya Demszky and Dana Movshovitz-Attias and Jeongwoo Ko and Alan S. Cowen and Gaurav Nemade and Sujith Ravi},
  booktitle={Annual Meeting of the Association for Computational Linguistics},
  year={2020},
  url={https://api.semanticscholar.org/CorpusID:218486942}
}

@article{Johnson2016MIMICIIIAF,
  title={MIMIC-III, a freely accessible critical care database},
  author={Alistair E. W. Johnson and Tom J. Pollard and Lu Shen and Li-wei H. Lehman and Mengling Feng and Mohammad Mahdi Ghassemi and Benjamin Moody and Peter Szolovits and Leo Anthony Celi and Roger G. Mark},
  journal={Scientific Data},
  year={2016},
  volume={3},
  url={https://api.semanticscholar.org/CorpusID:33285731}
}

@inproceedings{Mohammad2016SemEval2016T6,
  title={SemEval-2016 Task 6: Detecting Stance in Tweets},
  author={Saif M. Mohammad and Svetlana Kiritchenko and Parinaz Sobhani and Xiao-Dan Zhu and Colin Cherry},
  booktitle={International Workshop on Semantic Evaluation},
  year={2016},
  url={https://api.semanticscholar.org/CorpusID:286464}
}

@inproceedings{Margatina2021ActiveLB,
  title={Active Learning by Acquiring Contrastive Examples},
  author={Katerina Margatina and Giorgos Vernikos and Lo{\"i}c Barrault and Nikolaos Aletras},
  booktitle={Conference on Empirical Methods in Natural Language Processing},
  year={2021},
  url={https://api.semanticscholar.org/CorpusID:237442331}
}

@article{Sener2017ActiveLF,
  title={Active Learning for Convolutional Neural Networks: A Core-Set Approach},
  author={Ozan Sener and Silvio Savarese},
  journal={arXiv: Machine Learning},
  year={2017},
  url={https://api.semanticscholar.org/CorpusID:3383786}
}

@inproceedings{Xu2024EnergyBasedCB,
  title={Energy-Based Concept Bottleneck Models: Unifying Prediction, Concept Intervention, and Probabilistic Interpretations},
  author={Xinyue Xu and Yi Qin and Lu Mi and Hao Wang and Xiaomeng Li},
  booktitle={International Conference on Learning Representations},
  year={2024},
  url={https://api.semanticscholar.org/CorpusID:268031821}
}

@article{Hu2024SemiSupervisedCB,
  title={Semi-Supervised Concept Bottleneck Models},
  author={Lijie Hu and Tianhao Huang and Huanyi Xie and Chenyang Ren and Zhengyu Hu and Lu Yu and Di Wang},
  journal={2025 IEEE/CVF International Conference on Computer Vision (ICCV)},
  year={2024},
  pages={2110-2119},
  url={https://api.semanticscholar.org/CorpusID:270765040}
}

@article{Wei2022ChainOT,
  title={Chain of Thought Prompting Elicits Reasoning in Large Language Models},
  author={Jason Wei and Xuezhi Wang and Dale Schuurmans and Maarten Bosma and Ed H. Chi and F. Xia and Quoc Le and Denny Zhou},
  journal={ArXiv},
  year={2022},
  volume={abs/2201.11903},
  url={https://api.semanticscholar.org/CorpusID:246411621}
}

@article{Hsieh2023DistillingSO,
  title={Distilling Step-by-Step! Outperforming Larger Language Models with Less Training Data and Smaller Model Sizes},
  author={Cheng-Yu Hsieh and Chun-Liang Li and Chih-Kuan Yeh and Hootan Nakhost and Yasuhisa Fujii and Alexander J. Ratner and Ranjay Krishna and Chen-Yu Lee and Tomas Pfister},
  journal={ArXiv},
  year={2023},
  volume={abs/2305.02301},
  url={https://api.semanticscholar.org/CorpusID:258461606}
}

@article{Xu2025GraphCB,
  title={Graph Concept Bottleneck Models},
  author={Haotian Xu and Tsui-Wei Weng and Lam M. Nguyen and Tengfei Ma},
  journal={ArXiv},
  year={2025},
  volume={abs/2508.14255},
  url={https://api.semanticscholar.org/CorpusID:280691737}
}

@article{Barbiero2023RelationalCB,
  title={Relational Concept Bottleneck Models},
  author={Pietro Barbiero and Francesco Giannini and Gabriele Ciravegna and Michelangelo Diligenti and Giuseppe Marra},
  journal={Advances in Neural Information Processing Systems 37},
  year={2023},
  url={https://api.semanticscholar.org/CorpusID:261076254}
}

@inproceedings{Culotta2005ReducingLE,
  title={Reducing Labeling Effort for Structured Prediction Tasks},
  author={Aron Culotta and Andrew McCallum},
  booktitle={AAAI Conference on Artificial Intelligence},
  year={2005},
  url={https://api.semanticscholar.org/CorpusID:11583646}
}

@inproceedings{Yuan2020ColdstartAL,
  title={Cold-start Active Learning through Self-Supervised Language Modeling},
  author={Michelle Yuan and Hsuan-Tien Lin and Jordan Lee Boyd-Graber},
  booktitle={Conference on Empirical Methods in Natural Language Processing},
  year={2020},
  url={https://api.semanticscholar.org/CorpusID:224724415}
}

@article{Cruickshank2023PromptingAF,
  title={Prompting and Fine-Tuning Open-Sourced Large Language Models for Stance Classification},
  author={Iain J. Cruickshank and Lynnette Hui Xian Ng},
  journal={ACM Transactions on Intelligent Systems and Technology},
  year={2023},
  url={https://api.semanticscholar.org/CorpusID:262465334}
}

@article{Rezk2024LLMsFC,
  title={LLMs for clinical risk prediction},
  author={Mohamed Rezk and Patricia Cabanillas Silva and Fried-Michael Dahlweid},
  journal={ArXiv},
  year={2024},
  volume={abs/2409.10191},
  url={https://api.semanticscholar.org/CorpusID:272690300}
}

@article{Kojima2022LargeLM,
  title={Large Language Models are Zero-Shot Reasoners},
  author={Takeshi Kojima and Shixiang Shane Gu and Machel Reid and Yutaka Matsuo and Yusuke Iwasawa},
  journal={ArXiv},
  year={2022},
  volume={abs/2205.11916},
  url={https://api.semanticscholar.org/CorpusID:249017743}
}

@article{Naveed2023ACO,
  title={A Comprehensive Overview of Large Language Models},
  author={Humza Naveed and Asad Khan and Shi Qiu and Muhammad Saqib and Saeed Anwar and Muhammad Usman and Nick Barnes and Ajmal Saeed Mian},
  journal={ACM Transactions on Intelligent Systems and Technology},
  year={2023},
  volume={16},
  pages={1 - 72},
  url={https://api.semanticscholar.org/CorpusID:259847443}
}

@article{Erol2025CostofPassAE,
  title={Cost-of-Pass: An Economic Framework for Evaluating Language Models},
  author={Mehmet Hamza Erol and Batu El and Mirac Suzgun and Mert Y{\"u}ksekg{\"o}n{\"u}l and James Zou},
  journal={ArXiv},
  year={2025},
  volume={abs/2504.13359},
  url={https://api.semanticscholar.org/CorpusID:277940358}
}

@article{Sanh2019DistilBERTAD,
  title={DistilBERT, a distilled version of BERT: smaller, faster, cheaper and lighter},
  author={Victor Sanh and Lysandre Debut and Julien Chaumond and Thomas Wolf},
  journal={ArXiv},
  year={2019},
  volume={abs/1910.01108},
  url={https://api.semanticscholar.org/CorpusID:203626972}
}

@inproceedings{Jiao2019TinyBERTDB,
  title={TinyBERT: Distilling BERT for Natural Language Understanding},
  author={Xiaoqi Jiao and Yichun Yin and Lifeng Shang and Xin Jiang and Xiao Chen and Linlin Li and Fang Wang and Qun Liu},
  booktitle={Findings},
  year={2019},
  url={https://api.semanticscholar.org/CorpusID:202719327}
}

@inproceedings{Sun2020MobileBERTAC,
  title={MobileBERT: a Compact Task-Agnostic BERT for Resource-Limited Devices},
  author={Zhiqing Sun and Hongkun Yu and Xiaodan Song and Renjie Liu and Yiming Yang and Denny Zhou},
  booktitle={Annual Meeting of the Association for Computational Linguistics},
  year={2020},
  url={https://api.semanticscholar.org/CorpusID:215238853}
}

@article{Wan2023EfficientLL,
  title={Efficient Large Language Models: A Survey},
  author={Zhongwei Wan and Xin Wang and Che Liu and Samiul Alam and Yu Zheng and Jiachen Liu and Zhongnan Qu and Shen Yan and Yi Zhu and Quanlu Zhang and Mosharaf Chowdhury and Mi Zhang},
  journal={ArXiv},
  year={2023},
  volume={abs/2312.03863},
  url={https://api.semanticscholar.org/CorpusID:266044196}
}

@inproceedings{Ho2022LargeLM,
  title={Large Language Models Are Reasoning Teachers},
  author={Namgyu Ho and Laura Schmid and Se-Young Yun},
  booktitle={Annual Meeting of the Association for Computational Linguistics},
  year={2022},
  url={https://api.semanticscholar.org/CorpusID:254877399}
}

@inproceedings{Bhat2023LargeLM,
  title={Large Language Models As Annotators: A Preliminary Evaluation For Annotating Low-Resource Language Content},
  author={Savita Bhat and Vasudeva Varma},
  booktitle={EVAL4NLP},
  year={2023},
  url={https://api.semanticscholar.org/CorpusID:268064052}
}

@article{Xia2025FromST,
  title={From Selection to Generation: A Survey of LLM-based Active Learning},
  author={Yu Xia and Subhojyoti Mukherjee and Zhouhang Xie and Junda Wu and Xintong Li and Ryan Aponte and Hanjia Lyu and Joe Barrow and Hongjie Chen and Franck Dernoncourt and Branislav Kveton and Tong Yu and Ruiyi Zhang and Jiuxiang Gu and Nesreen K. Ahmed and Yu Wang and Xiang Chen and Hanieh Deilamsalehy and Sungchul Kim and Zhengmian Hu and Yue Zhao and Nedim Lipka and Seunghyun Yoon and Tinghui Huang and Zichao Wang and Puneet Mathur and Soumyabrata Pal and Koyel Mukherjee and Zhehao Zhang and Namyong Park and Thien Huu Nguyen and Jiebo Luo and Ryan A. Rossi and Julian J. McAuley},
  journal={ArXiv},
  year={2025},
  volume={abs/2502.11767},
  url={https://api.semanticscholar.org/CorpusID:276408942}
}

@article{Xiang2025PromptALSD,
  title={PromptAL: Sample-Aware Dynamic Soft Prompts for Few-Shot Active Learning},
  author={Hui Xiang and Jinqiao Shi and Ting Zhang and Xiaojie Zhao and Yong Liu and Yong Ma},
  journal={ArXiv},
  year={2025},
  volume={abs/2507.16424},
  url={https://api.semanticscholar.org/CorpusID:280166820}
}

@inproceedings{Shridhar2022DistillingRC,
  title={Distilling Reasoning Capabilities into Smaller Language Models},
  author={Kumar Shridhar and Alessandro Stolfo and Mrinmaya Sachan},
  booktitle={Annual Meeting of the Association for Computational Linguistics},
  year={2022},
  url={https://api.semanticscholar.org/CorpusID:258762841}
}

@article{Yao2023TreeOT,
  title={Tree of Thoughts: Deliberate Problem Solving with Large Language Models},
  author={Shunyu Yao and Dian Yu and Jeffrey Zhao and Izhak Shafran and Thomas L. Griffiths and Yuan Cao and Karthik Narasimhan},
  journal={ArXiv},
  year={2023},
  volume={abs/2305.10601},
  url={https://api.semanticscholar.org/CorpusID:258762525}
}

@article{Yao2023BeyondCE,
  title={Beyond Chain-of-Thought, Effective Graph-of-Thought Reasoning in Language Models},
  author={Yao Yao and Z. Li and Hai Zhao},
  journal={ArXiv},
  volume={abs/2305.16582},
  year={2023},
  url={https://arxiv.org/abs/2305.16582}
}

@article{Yang2023LargeLM,
  title={Large Language Models as Optimizers},
  author={Chengrun Yang and Xuezhi Wang and Yifeng Lu and Hanxiao Liu and Quoc V. Le and Denny Zhou and Xinyun Chen},
  journal={ArXiv},
  year={2023},
  volume={abs/2309.03409},
  url={https://api.semanticscholar.org/CorpusID:261582296}
}

@article{Huang2022TowardsRI,
  title={Towards Reasoning in Large Language Models: A Survey},
  author={Jie Huang and Kevin Chen-Chuan Chang},
  journal={ArXiv},
  year={2022},
  volume={abs/2212.10403},
  url={https://api.semanticscholar.org/CorpusID:254877753}
}

@article{Dubey2024TheL3,
  title={The Llama 3 Herd of Models},
  author={Abhimanyu Dubey and Abhinav Jauhri and Abhinav Pandey and Abhishek Kadian and Ahmad Al-Dahle and Aiesha Letman and Akhil Mathur and Alan Schelten and Amy Yang and Angela Fan and Anirudh Goyal and Anthony S. Hartshorn and Aobo Yang and Archi Mitra and Archie Sravankumar and Artem Korenev and Arthur Hinsvark and Arun Rao and Aston Zhang and Aur'elien Rodriguez and Austen Gregerson and Ava Spataru and Baptiste Rozi{\`e}re and Bethany M. Biron and Binh Tang and Bobbie Chern and Charlotte Caucheteux and Chaya Nayak and Chloe Bi and Chris Marra and Chris McConnell and Christian Keller and Christophe Touret and Chunyang Wu and Corinne Wong and Cristian Canton Ferrer and Cyrus Nikolaidis and Damien Allonsius and Daniel J. Song and Danielle Pintz and Danny Livshits and David Esiobu and Dhruv Choudhary and Dhruv Mahajan and Diego Garcia-Olano and Diego Perino and Dieuwke Hupkes and Egor Lakomkin and Ehab A. AlBadawy and E I Lobanova and Emily Dinan and Eric Michael Smith and Filip Radenovic and Frank Zhang and Gabriel Synnaeve and Gabrielle Lee and Georgia Lewis Anderson and Graeme Nail and Gr{\'e}goire Mialon and Guanglong Pang and Guillem Cucurell and Hailey Nguyen and Hannah Korevaar and Hu Xu and Hugo Touvron and Iliyan Zarov and Imanol Arrieta Ibarra and Isabel M. Kloumann and Ishan Misra and Ivan Evtimov and Jade Copet and Jaewon Lee and Jan Geffert and Jana Vranes and Jason Park and Jay Mahadeokar and Jeet Shah and Jelmer van der Linde and Jennifer Billock and Jenny Hong and Jenya Lee and Jeremy Fu and Jianfeng Chi and Jianyu Huang and Jiawen Liu and Jie Wang and Jiecao Yu and Joanna Bitton and Joe Spisak and Jongsoo Park and Joseph Rocca and Joshua Johnstun and Joshua Saxe and Ju-Qing Jia and Kalyan Vasuden Alwala and K. Upasani and Kate Plawiak and Keqian Li and Kenneth Heafield and Kevin R. Stone and Khalid El-Arini and Krithika Iyer and Kshitiz Malik and Kuen-ley Chiu and Kunal Bhalla and Lauren Rantala-Yeary and Laurens van der Maaten and Lawrence Chen and Liang Tan and Liz Jenkins and Louis Martin and Lovish Madaan and Lubo Malo and Lukas Blecher and Lukas Landzaat and Luke de Oliveira and Madeline Muzzi and Ma-hesh Pasupuleti and Mannat Singh and Manohar Paluri and Marcin Kardas and Mathew Oldham and Mathieu Rita and Maya Pavlova and Melissa Hall Melanie Kambadur and Mike Lewis and Min Si and Mitesh Kumar Singh and Mona Hassan and Naman Goyal and Narjes Torabi and Niko-lay Bashlykov and Nikolay Bogoychev and Niladri S. Chatterji and Olivier Duchenne and Onur cCelebi and Patrick Alrassy and Pengchuan Zhang and Pengwei Li and Petar Vasi{\'c} and Peter Weng and Prajjwal Bhargava and Pratik Dubal and Praveen Krishnan and Punit Singh Koura and Puxin Xu and Qing He and Qingxiao Dong and Ragavan Srinivasan and Raj Ganapathy and Ramon Calderer and Ricardo Silveira Cabral and Robert Stojnic and Roberta Raileanu and Rohit Girdhar and Rohit Patel and Romain Sauvestre and Ron-nie Polidoro and Roshan Sumbaly and Ross Taylor and Ruan Silva and Rui Hou and Rui Wang and Saghar Hosseini and Sa-hana Chennabasappa and Sanjay Singh and Sean Bell and Seohyun Sonia Kim and Sergey Edunov and Shaoliang Nie and Sharan Narang and Sharath Chandra Raparthy and Sheng Shen and Shengye Wan and Shruti Bhosale and Shun Zhang and Simon Vandenhende and Soumya Batra and Spencer Whitman and Sten Sootla and St{\'e}phane Collot and Suchin Gururangan and Sydney Borodinsky and Tamar Herman and Tara Fowler and Tarek Sheasha and Thomas Georgiou and Thomas Scialom and Tobias Speckbacher and Todor Mihaylov and Tong Xiao and Ujjwal Karn and Vedanuj Goswami and Vibhor Gupta and Vignesh Ramanathan and Viktor Kerkez and Vincent Gonguet and Virginie Do and Vish Vogeti and Vladan Petrovic and Weiwei Chu and Wenhan Xiong and Wenyin Fu and Whit-ney Meers and Xavier Martinet and Xiaodong Wang and Xiaoqing Ellen Tan and Xinfeng Xie and Xuchao Jia and Xuewei Wang and Yaelle Goldschlag and Yashesh Gaur and Yasmine Babaei and Yiqian Wen and Yiwen Song and Yuchen Zhang and Yue Li and Yuning Mao and Zacharie Delpierre Coudert and Zhengxu Yan and Zhengxing Chen and Zoe Papakipos and Aaditya K. Singh and Aaron Grattafiori and Abha Jain and Adam Kelsey and Adam Shajnfeld and Adi Gangidi and Adolfo Victoria and Ahuva Goldstand and Ajay Menon and Ajay Sharma and Alex Boesenberg and Alex Vaughan and Alexei Baevski and Allie Feinstein and Amanda Kallet and Amit Sangani and Anam Yunus and Andrei Lupu and Andres Alvarado and Andrew Caples and Andrew Gu and Andrew Ho and Andrew Poulton and Andrew Ryan and Ankit Ramchandani and Annie Franco and Aparajita Saraf and Arkabandhu Chowdhury and Ashley Gabriel and Ashwin Bharambe and Assaf Eisenman and Azadeh Yazdan and Beau James and Ben Maurer and Benjamin Leonhardi and Po-Yao (Bernie) Huang and Beth Loyd and Beto de Paola and Bhargavi Paranjape and Bing Liu and Bo Wu and Boyu Ni and Braden Hancock and Bram Wasti and Brandon Spence and Brani Stojkovic and Brian Gamido and Britt Montalvo and Carl Parker and Carly Burton and Catalina Mejia and Changhan Wang and Changkyu Kim and Chao Zhou and Chester Hu and Ching-Hsiang Chu and Chris Cai and Chris Tindal and Christoph Feichtenhofer and Damon Civin and Dana Beaty and Daniel Kreymer and Shang-Wen Li and Danny Wyatt and David Adkins and David Xu and Davide Testuggine and Delia David and Devi Parikh and Diana Liskovich and Didem Foss and Dingkang Wang and Duc Le and Dustin Holland and Edward Dowling and Eissa Jamil and Elaine Montgomery and Eleonora Presani and Emily Hahn and Emily Wood and Erik Brinkman and Esteban Arcaute and Evan Dunbar and Evan Smothers and Fei Sun and Felix Kreuk and Feng Tian and Firat Ozgenel and Francesco Caggioni and Francisco (Paco) Guzm{\'a}n and Frank J. Kanayet and Frank Seide and Gabriela Medina Florez and Gabriella Schwarz and Gada Badeer and Georgia Swee and Gil Halpern and Govind Thattai and Grant Herman and Grigory Sizov and Guangyi Zhang and Guna Lakshminarayanan and Hamid Shojanazeri and Han Zou and Hannah Wang and Han Zha and Haroun Habeeb and Harrison Rudolph and Helen Suk and Henry Aspegren and Hunter Goldman and Igor Molybog and Igor Tufanov and Irina-Elena Veliche and Itai Gat and Jake Weissman and James Geboski and James Kohli and Japhet Asher and Jean-Baptiste Gaya and Jeff Marcus and Jeff Tang and Jennifer Chan and Jenny Zhen and Jeremy Reizenstein and Jeremy Teboul and Jessica Zhong and Jian Jin and Jingyi Yang and Joe Cummings and Jon Carvill and Jon Shepard and Jonathan McPhie and Jonathan Torres and Josh Ginsburg and Junjie Wang and Kaixing(Kai) Wu and U KamHou and Karan Saxena and Karthik Prasad and Kartikay Khandelwal and Katayoun Zand and Kathy Matosich and Kaushik Veeraraghavan and Kelly Michelena and Keqian Li and Kun Huang and Kunal Chawla and Kushal Lakhotia and Kyle Huang and Lailin Chen and Lakshya Garg and A Lavender and Leandro Silva and Lee Bell and Lei Zhang and Liangpeng Guo and Licheng Yu and Liron Moshkovich and Luca Wehrstedt and Madian Khabsa and Manav Avalani and Manish Bhatt and Maria Tsimpoukelli and Martynas Mankus and Matan Hasson and Matthias Lennie and Matthias Reso and Maxim Groshev and Maxim Naumov and Maya Lathi and Meghan Keneally and Michael L. Seltzer and Michal Valko and Michelle Restrepo and Mihir Patel and Mik Vyatskov and Mikayel Samvelyan and Mike Clark and Mike Macey and Mike Wang and Miquel Jubert Hermoso and Mo Metanat and Mohammad Rastegari and Mun-ish Bansal and Nandhini Santhanam and Natascha Parks and Natasha White and Navy-ata Bawa and Nayan Singhal and Nick Egebo and Nicolas Usunier and Nikolay Pavlovich Laptev and Ning Dong and Ning Zhang and Norman Cheng and Oleg Chernoguz and Olivia Hart and Omkar Salpekar and Ozlem Kalinli and Parkin Kent and Parth Parekh and Paul Saab and Pavan Balaji and Pe-dro Rittner and Philip Bontrager and Pierre Roux and Piotr Doll{\'a}r and Polina Zvyagina and Prashant Ratanchandani and Pritish Yuvraj and Qian Liang and Rachad Alao and Rachel Rodriguez and Rafi Ayub and Raghotham Murthy and Raghu Nayani and Rahul Mitra and Raymond Li and Rebekkah Hogan and Robin Battey and Rocky Wang and Ro-han Maheswari and Russ Howes and Ruty Rinott and Sai Jayesh Bondu and Samyak Datta and Sara Chugh and Sara Hunt and Sargun Dhillon and S. Yu. Sidorov and Satadru Pan and Saurabh Verma and Seiji Yamamoto and Sharadh Ramaswamy and Shaun Lindsay and Sheng Feng and Shenghao Lin and Shengxin Zha and Shiva Shankar and Shuqiang Zhang and Sinong Wang and Sneha Agarwal and Soji Sajuyigbe and Soumith Chintala and Stephanie Max and Stephen Chen and Steve Kehoe and Steve Satterfield and Sudarshan Govindaprasad and Sumit Kumar Gupta and Sung-Bae Cho and Sunny Virk and Suraj Subramanian and Sy Choudhury and Sydney Goldman and Tal Remez and Tamar Glaser and Tamara Best and Thilo Kohler and Thomas Robinson and Tianhe Li and Tianjun Zhang and Tim Matthews and Timothy Chou and Tzook Shaked and Varun Vontimitta and Victoria O Ajayi and Victoria Montanez and Vijai Mohan and Vinay Kumar and Vishal Mangla and Vlad Ionescu and Vlad Andrei Poenaru and Vlad T. Mihailescu and Vladimir Ivanov and Wei Li and Wenchen Wang and Wenwen Jiang and Wes Bouaziz and Will Constable and Xia Tang and Xiaofang Wang and Xiaojian Wu and Xiaolan Wang and Xide Xia and Xilun Wu and Xinbo Gao and Yanjun Chen and Ye Hu and Ye Jia and Ye Qi and Yenda Li and Yilin Zhang and Ying Zhang and Yossi Adi and Youngjin Nam and Yu Wang and Yuchen Hao and Yundi Qian and Yuzi He and Zach Rait and Zachary DeVito and Zef Rosnbrick and Zhaoduo Wen and Zhenyu Yang and Zhiwei Zhao},
  journal={ArXiv},
  volume={abs/2407.21783},
  year={2024},
  url={https://arxiv.org/abs/2407.21783}
}

@inproceedings{karimi2016linear,
  title={Linear Convergence of Gradient and Proximal-Gradient Methods Under the Polyak-{\L}ojasiewicz Condition},
  author={Hamed Karimi and Julie Nutini and Mark Schmidt},
  booktitle={Machine Learning and Knowledge Discovery in Databases (ECML PKDD)},
  pages={795--811},
  year={2016},
  url={https://api.semanticscholar.org/CorpusID:13116679}
}

@techreport{Settles2009ActiveLearning,
  title={Active Learning Literature Survey},
  author={Burr Settles},
  institution={University of Wisconsin--Madison Department of Computer Sciences},
  number={1648},
  year={2009},
  url={https://minds.wisconsin.edu/handle/1793/60660}
}

@misc{reimers2019sentencebertsentenceembeddingsusing,
      title={Sentence-BERT: Sentence Embeddings using Siamese BERT-Networks}, 
      author={Nils Reimers and Iryna Gurevych},
      year={2019},
      eprint={1908.10084},
      archivePrefix={arXiv},
      primaryClass={cs.CL},
      url={https://arxiv.org/abs/1908.10084}, 
}

\clearpage
\appendix

\section{End-to-End Algorithm}
\label{sec:alg_gcp}

Algorithm~\ref{alg:gcp} states the complete GCP loop of
Section~\ref{sec:execution}. Line~1 builds the concept graph once and
initializes $\mathcal{D}$ with the annotations of $\mathcal{X}_0$, the same
annotations that selected the concepts and edges
(Section~\ref{sec:graph_construction}), and Line~2 trains every predictor
of the initial student with Eq.~\eqref{eq:training}; no later line updates
all predictors. Each round (Lines~3--10) follows
Figure~\ref{fig:framework}. Lines~4--5 compute the node-level outputs on
$\mathcal{U}$ and select $\mathcal{S}^\ast$ by graph-aware acquisition
(Section~\ref{sec:acquisition}). Line~6 queries the teacher once per
selected text for its label and all concept values, and Line~7 moves the
annotated texts into $\mathcal{D}$. Lines~8--9 compute the influence scores
on the enlarged $\mathcal{D}$ and retrain only the top-$b$ predictors, on
$\mathcal{D}$ and generated texts, while the other predictors stay frozen
(Section~\ref{sec:submodule} and Algorithm~\ref{alg:submodule_retrain}).
The loop ends once the teacher tokens spent in Lines~1, 6, and~9 reach the
budget $N$ (Line~3).

\begin{algorithm}[ht]
\small
\caption{GCP active distillation}
\label{alg:gcp}
\begin{algorithmic}[1]
\STATE Build $G$ (\S\ref{sec:graph_construction}); $\mathcal{D}\leftarrow$ annotated $\mathcal{X}_0$
\STATE Train $\{f_j\}_{j\in V}$ on $\mathcal{D}$ by minimizing Eq.~\eqref{eq:training}
\WHILE{teacher tokens spent $<N$}
  \STATE Compute $p_i(x)$ and $z_i(x)$ for all $x\in\mathcal{U}$, $i\in V$
  \STATE $\mathcal{S}^\ast\leftarrow$ consensus selection of $B$ texts (\S\ref{sec:acquisition})
  \STATE Query the teacher for $(\mathbf{c},y)$ of every $x\in\mathcal{S}^\ast$
  \STATE $\mathcal{D}\leftarrow\mathcal{D}\cup\mathcal{S}^\ast$;\ \ $\mathcal{U}\leftarrow\mathcal{U}\setminus\mathcal{S}^\ast$
  \STATE Compute $\Delta_i$ on $\mathcal{D}$ and $\mathcal{M}^\star$ (\S\ref{sec:submodule})
  \STATE Retrain $f_i$, $i\in\mathcal{M}^\star$, on $\mathcal{D}$ and generated texts; freeze $f_j$, $j\notin\mathcal{M}^\star$
\ENDWHILE
\STATE \textbf{return} $\{f_j\}_{j\in V}$
\end{algorithmic}
\end{algorithm}

\section{Sub-Module Retraining Algorithm}
\label{sec:algorithm}

Algorithm~\ref{alg:submodule_retrain} details the targeted sub-module
retraining procedure introduced in Section~\ref{sec:submodule}. For each
annotated example, we estimate a node-level influence score $\Delta_i$ by
comparing the final loss under counterfactual reruns that include versus
exclude node $i$'s prediction. We then retrain only the top-$b$ most
influential predictors, each on its node loss over $\mathcal{D}$ and, for a
concept node, on LLM-generated texts labeled with its parent and child
concepts only; all other predictors stay frozen.

\begin{algorithm}[ht]
\small
\caption{Sub-module Retraining via Counterfactual Reruns}
\label{alg:submodule_retrain}
\begin{algorithmic}
\STATE {\bfseries Input:} GCP DAG $\mathcal{G}=(\mathcal{V},\mathcal{E})$ with predictors $\{f_i\}$;
annotated dataset $\mathcal{D}=\{(x,\mathbf{h},y)\}$;
final loss $\mathcal{L}$; retrain budget $b$
\STATE {\bfseries Output:} Updated predictors $\{f_i\}_{i\in\mathcal{V}}$

\STATE Initialize $\Delta_i \leftarrow 0$ for all $i\in\mathcal{V}$

\FOR{each $(x,\mathbf{h},y)\in\mathcal{D}$}
    \FOR{each node $i\in\mathcal{V}$}
        \STATE $\tilde{y}^{\mathrm{pa}}_i \leftarrow
        \mathrm{Rerun}(x;\mathrm{pa}(i),\mathbf{h}_{\mathrm{pa}(i)})$
        \STATE $\tilde{y}^{\mathrm{pa}+i}_i \leftarrow
        \mathrm{Rerun}(x;\mathrm{pa}(i)\cup\{i\},
        \mathbf{h}_{\mathrm{pa}(i)}\cup\{h_i\})$
        \STATE $\Delta_i \leftarrow \Delta_i +
        \mathcal{L}(\tilde{y}^{\mathrm{pa}}_i,y) -
        \mathcal{L}(\tilde{y}^{\mathrm{pa}+i}_i,y)$
    \ENDFOR
\ENDFOR

\STATE $\Delta_i \leftarrow \Delta_i / |\mathcal{D}|$ for all $i\in\mathcal{V}$

\STATE Select $\mathcal{M}_{\mathrm{retrain}} \leftarrow
\mathrm{Top}(\{\Delta_i\}_{i\in\mathcal{V}}, b)$ and freeze $f_j$ for $j \notin \mathcal{M}_{\mathrm{retrain}}$

\FOR{each $i\in\mathcal{M}_{\mathrm{retrain}}$}
    \STATE $\mathcal{D}_i \leftarrow \mathcal{D}$
    \IF{$i$ is a concept node}
        \STATE Generate short texts $x'$ labeled with parent and child concepts $(x',\mathbf{h}_{\mathrm{pa}(i)},h_i)$ using LLM; add them to $\mathcal{D}_i$
    \ENDIF
    \STATE Retrain $f_i$ on $\mathcal{D}_i$ by minimizing $F_i$, starting from its current weights
\ENDFOR
\end{algorithmic}
\end{algorithm}

\section{Experiment Details}
\label{sec:experiment details}

\subsection{Datasets Details}
\label{sec:Datasets Details}

We provide detailed descriptions and sources for the eight datasets used in
our experiments:

\begin{itemize}[leftmargin=*]
    \item \textbf{AG News}~\cite{Zhang2015CharacterlevelCN} is a topic
    classification dataset constructed from news articles in the AG corpus.
    Each instance contains a headline and short description and is assigned
    to one of four categories: \emph{World}, \emph{Sports},
    \emph{Business}, or \emph{Science/Technology}.

    \item \textbf{Amazon Review Polarity}~\cite{Zhang2015CharacterlevelCN}
    is built from user-written Amazon product reviews. Reviews rated 1--2
    stars are labeled \emph{negative}, those rated 4--5 stars
    \emph{positive}, and 3-star reviews are excluded. The rating is used
    only to derive the label; the model sees only the review text.
    
    \item \textbf{IMDB Movie Reviews}~\cite{Maas2011LearningWV} is a
    benchmark sentiment dataset of user-written movie reviews. Each review
    is classified as \emph{positive} or \emph{negative}.
    
    \item \textbf{Yelp Review Polarity}~\cite{Zhang2015CharacterlevelCN}
    contains user reviews of businesses from the Yelp Dataset Challenge.
    Each review is labeled \emph{positive} or \emph{negative} according to
    its star rating, and the model sees only the review text.

    \item \textbf{MNLI (Multi-Genre Natural Language Inference)}
    ~\cite{Williams2017ABC} contains premise--hypothesis pairs spanning
    multiple text genres. The task classifies each pair as
    \emph{entailment}, \emph{contradiction}, or \emph{neutral}.

    \item \textbf{GoEmotions}~\cite{Demszky2020GoEmotionsAD} consists of
    short, user-generated texts annotated with 27 fine-grained emotion
    categories. It is a multi-label task with substantial subjectivity and
    label overlap.

    \item \textbf{SemEval Stance Detection}
    ~\cite{Mohammad2016SemEval2016T6} pairs each text with a target entity or
    topic. The task identifies whether the stance toward that target is
    \emph{favor}, \emph{against}, or \emph{neutral}.

    \item \textbf{MIMIC-III Clinical Notes}~\cite{Johnson2016MIMICIIIAF} is a
    large clinical corpus derived from intensive-care electronic health
    records. We use patient-level notes, such as discharge summaries, to
    predict clinical outcomes or phenotypes including in-hospital mortality.

\end{itemize}

\begin{table}[ht]
\centering
\caption{Summary of dataset statistics. \# Instances is the size of the full
dataset, with all splits combined.}
\label{tab:dataset_statistics}
\tiny
\setlength{\tabcolsep}{3pt}
\resizebox{\columnwidth}{!}{%
\begin{tabular}{lcccc}
\toprule
Dataset & \# Instances & \# Classes & Avg. Length & Task Type \\
\midrule
AG News & 127,600 & 4 & 38 & Topic \\
Amazon Review Polarity & 4,000,000 & 2 & 90 & Sentiment \\
IMDB & 50,000 & 2 & 230 & Sentiment \\
Yelp Review Polarity & 598,000 & 2 & 150 & Sentiment \\
MNLI & 433,000 & 3 & 33 & NLI \\
GoEmotions & 58,000 & 27 & 30 & Emotion \\
SemEval Stance & 4,800 & 3 & 25 & Stance \\
MIMIC-III (In-hospital Mortality) & 50,000 & 2 & 1,200 & Clinical \\
\bottomrule
\end{tabular}%
}
\end{table}

Table~\ref{tab:dataset_statistics} summarizes the dataset sizes, label
counts, average lengths, and task types.

\subsection{Evaluation Metric} 
We evaluate the student model using test-set accuracy. For the single-label
datasets (MNLI, SemEval, AG News, Amazon, IMDB, and Yelp), accuracy is the
fraction of examples whose predicted label matches the ground truth. For
GoEmotions, we binarize each label and report example-level exact-match
accuracy over the predicted label set. For MIMIC-III, we report patient-level
accuracy for the downstream clinical endpoint.

\subsection{Baseline Methods}

We compare GCP against widely used active-learning acquisition strategies.
The uncertainty-based group includes Random sampling, Least Confidence
~\cite{Culotta2005ReducingLE}, Entropy Sampling
~\cite{Settles2009ActiveLearning}, and Disagreement (Vote Entropy)
~\cite{Siddhant2018DeepBA}. Stronger diversity-, representation-, and
gradient-based baselines include CoreSet (k-center)
~\cite{Sener2017ActiveLF}, BADGE~\cite{Ash2019DeepBA}, ALPS
~\cite{Yuan2020ColdstartAL}, and CAL~\cite{Margatina2021ActiveLB}. These
methods use prediction uncertainty, embedding diversity, or gradient
information, but none exploits explicit concept structure or reasoning
dependencies.

\subsection{Implementation Details}

We use LLaMA-3-70B as the annotator~\cite{Dubey2024TheL3}. All models are
implemented in PyTorch, and experiments run on a single NVIDIA H200 Tensor
Core GPU with 141GB HBM3e memory. We optimize with AdamW using a learning
rate of $1\times10^{-4}$ and batch size 8, applying gradient accumulation
when necessary to maintain stable updates. Unless otherwise stated, each
concept predictor is a lightweight MLP with hidden dimension 256 and dropout
0.1. Training hyperparameters remain fixed across active-learning rounds and
annotation budgets.

\section{Theoretical Analysis}
\label{sec:proof}

This appendix provides the complete definitions, assumptions, formal
statements, scope qualifications, and proofs for the theoretical results
stated in Section~\ref{sec:analysis}. Each result is restated under its
main-text number.

\subsection{Formal Statements and Scope}

Because GCP is supervised at every concept node
(Section~\ref{sec:submodule}), the comparison is stated for the joint
concept--label risk that the model is trained to minimize.

\begin{definition}[Joint concept--label risk]
\label{def:joint_risk}
For a model class $\mathcal{P}$ whose members specify a conditional
$p(y,\mathbf{c}\mid x)$ over the label and all concepts, define
\[
\mathcal{R}(\mathcal{P})
=
\inf_{p\in\mathcal{P}}
\mathbb{E}_{p^\star}\big[-\log p(y,\mathbf{c}\mid x)\big].
\]
\end{definition}

\begin{restated}{Theorem~\ref{thm:gcp_risk}}[Joint-risk advantage of GCP]
Let $G=(V,E)$ be a DAG over $V=\{c_1,\ldots,c_K,y\}$ and assume the
data-generating distribution factorizes as
\[
\begin{aligned}
p^\star(y,\mathbf{c}\mid x)
&=
\prod_{j=1}^K
p^\star\!\big(c_j\mid x,\mathbf{c}_{\mathrm{pa}^\star(c_j)}\big)\\[-1mm]
&\quad{}\times
p^\star\!\big(y\mid x,\mathbf{c}_{\mathrm{pa}^\star(y)}\big).
\end{aligned}
\]
Define
\begin{align*}
\mathcal{P}_{\mathrm{GCP}}
&=\{p:\ p \text{ is DAG-factorized on } V\},\\[-1mm]
\mathcal{P}_{\mathrm{CBM}}
&=\left\{p:\begin{aligned}
p(y,\mathbf{c}\mid x)&=p(y\mid x,\mathbf{c})\\[-1mm]
&\quad{}\times\textstyle\prod_jp(c_j\mid x)
\end{aligned}\right\},\\[-1mm]
\mathcal{P}_{\mathrm{MLP}}
&=\left\{p:\begin{aligned}
p(y,\mathbf{c}\mid x)&=p(y\mid x)\\[-1mm]
&\quad{}\times\textstyle\prod_jp(c_j\mid x)
\end{aligned}\right\}.
\end{align*}
Then
$\mathcal{P}_{\mathrm{MLP}}\subseteq\mathcal{P}_{\mathrm{CBM}}
\subseteq\mathcal{P}_{\mathrm{GCP}}$, and
\[
\mathcal{R}(\mathcal{P}_{\mathrm{CBM}})
-\mathcal{R}(\mathcal{P}_{\mathrm{GCP}})
=
\mathbb{E}_{p^\star(x)}\!\big[\mathrm{TC}(\mathbf{c}\mid x)\big],
\]
\[
\mathcal{R}(\mathcal{P}_{\mathrm{MLP}})
-\mathcal{R}(\mathcal{P}_{\mathrm{CBM}})
=
\mathbb{E}_{p^\star(x)}\!\big[I(y;\mathbf{c}\mid x)\big],
\]
where
$\mathrm{TC}(\mathbf{c}\mid x)
=\mathrm{KL}\big(p^\star(\mathbf{c}\mid x)
\big\|\prod_jp^\star(c_j\mid x)\big)$.
The first inequality is strict whenever some
$c_i\not\!\perp\!\!\!\perp c_j\mid x$ on a set of positive probability;
the second is strict whenever
$y\not\!\perp\!\!\!\perp\mathbf{c}\mid x$.
\end{restated}

\begin{remark}[Joint risk versus label-only risk]
\label{rem:joint_vs_label}
Theorem~\ref{thm:gcp_risk} is not a population label-only risk claim.
An unconstrained predictor can represent any label marginal, including
one induced by a GCP, so its population label-only risk can coincide
with that of GCP. The theorem instead quantifies the excess paid by flat
and unstructured classes when fitting the supervised concept--label
joint. The label-accuracy gap in
Table~\ref{tab:ablation-architecture} is therefore a finite-capacity,
finite-sample effect in which the correct factorization acts as a
structural regularizer.
\end{remark}

The convergence result concerns the teacher-forced training objective:
annotated rather than predicted parent concepts enter each node loss.

\begin{assumption}[Teacher-forced training objective]
\label{asm:teacher_forcing}
During training,
$F_v(\theta_v)=
\mathbb{E}[-\log p(v\mid x,\mathbf{c}_{\mathrm{pa}(v)};\theta_v)]$
uses annotated parent concepts, so $F_v$ depends on $\theta_v$ alone.
\end{assumption}

\begin{restated}{Theorem~\ref{thm:gcp_conv}}[Linear convergence of GCP]
Let
$F(\theta)=\sum_{v\in V}\lambda_vF_v(\theta_v)$ with $\lambda_v>0$.
Under Assumption~\ref{asm:teacher_forcing}, suppose each $F_v$ is
$L_v$-smooth and satisfies a local Polyak--\L{}ojasiewicz inequality
\citep{karimi2016linear} with constant $\mu_v$ near a minimizer.
Then $F$ is $L$-smooth for $L=\max_v\lambda_vL_v$, satisfies a local
PL inequality with
\[
\mu_{\mathrm{GCP}}=\min_{v\in V}\lambda_v\mu_v,
\]
and gradient descent with step size $1/L$ obeys
\[
F(\theta_t)-F(\theta^\star)
\le
\left(1-\frac{\mu_{\mathrm{GCP}}}{L}\right)^t
\big(F(\theta_0)-F(\theta^\star)\big).
\]
Parameter-distance contraction additionally requires quadratic growth
on the same neighbourhood.
\end{restated}

\begin{assumption}[Curvature monotonicity under conditioning]
\label{asm:curvature}
For each node $v$, the local PL constant does not decrease when the node
conditions on additional informative parents:
$\mu_v^{\mathrm{GCP}}\ge\mu_v^{\mathrm{CBM}}
\ge\mu_v^{\mathrm{MLP}}$.
\end{assumption}

\begin{restated}{Proposition~\ref{prop:mu_order}}[Ordering of convergence constants]
Suppose Assumptions~\ref{asm:teacher_forcing}
and~\ref{asm:curvature} hold with the same positive weights across model
classes. Then
$\mu_{\mathrm{GCP}}\ge\mu_{\mathrm{CBM}}\ge\mu_{\mathrm{MLP}}$.
If both node-level comparisons are strict for every node, the global
comparisons are strict and GCP has the faster linear rate in
Theorem~\ref{thm:gcp_conv}.
\end{restated}

\begin{remark}[What is assumed and what is proved]
\label{rem:conv_scope}
Theorem~\ref{thm:gcp_conv} follows from teacher forcing and node-level
regularity. Proposition~\ref{prop:mu_order} is conditional on
Assumption~\ref{asm:curvature}, which is not derived. Because the global
constant is a minimum over nodes, improvement at only some nodes need
not increase it. Moreover, separability holds for the teacher-forced
training objective, not for inference with predicted parents.
\end{remark}

\begin{restated}{Theorem~\ref{thm:optimal}}[Optimal retraining selection]
For empirical counterfactual scores
\[
\widehat{\Delta}_i
=
\frac{1}{|\mathcal{D}|}
\sum_{(x,\mathbf{h},y)\in\mathcal{D}}
\big(
\mathcal{L}^{\mathrm{pa}}_i(x)
-\mathcal{L}^{\mathrm{pa}+i}_i(x)
\big),
\]
the additive selection problem
\begin{equation}
\max_{S\subseteq V,\,|S|=b}
\sum_{i\in S}\widehat{\Delta}_i
\label{eq:obj}
\end{equation}
is solved globally by selecting the $b$ largest scores. Moreover, for
$|\mathcal{D}|$ annotated examples and a graph with $|V|$ nodes, dense
counterfactual sub-module retraining has worst-case complexity
$\mathcal{O}(|\mathcal{D}||V|^2)$.
\end{restated}

\subsection{Proof of Theorem~\ref{thm:gcp_risk}}

\label{sec:proof_gcp_risk}

Throughout, all risks are the joint risk of
Definition~\ref{def:joint_risk}. The argument is carried out entirely on
the joint distribution $(y,\mathbf{c})$; no step marginalizes over
$\mathbf{c}$.

\paragraph{Step 0: risk as an excess KL.}
For any $p$ specifying $p(y,\mathbf{c}\mid x)$,
\begin{align*}
&\mathbb{E}_{p^\star}\big[-\log p(y,\mathbf{c}\mid x)\big]\\
&\quad=H^\star+
\mathbb{E}_{p^\star(x)}\!\Big[\\[-1mm]
&\qquad
\mathrm{KL}\big(
p^\star(y,\mathbf{c}\mid x)
\,\|\,p(y,\mathbf{c}\mid x)
\big)\Big],
\end{align*}
where $H^\star:=\mathbb{E}_{p^\star}[-\log p^\star(y,\mathbf{c}\mid x)]$ is the
Bayes joint risk and does not depend on $p$. Minimizing
$\mathcal{R}$ over a class is therefore equivalent to minimizing the
expected KL term over that class.

\paragraph{Step 1: nesting of the classes.}
Given $p\in\mathcal{P}_{\mathrm{MLP}}$, setting $p(y\mid x,\mathbf{c})\equiv p(y\mid x)$
exhibits it as an element of $\mathcal{P}_{\mathrm{CBM}}$, so
$\mathcal{P}_{\mathrm{MLP}}\subseteq\mathcal{P}_{\mathrm{CBM}}$. A flat CBM is the
DAG factorization with $\mathrm{pa}(c_j)=\varnothing$ for all $j$ and
$\mathrm{pa}(y)=\{c_1,\ldots,c_K\}$, so
$\mathcal{P}_{\mathrm{CBM}}\subseteq\mathcal{P}_{\mathrm{GCP}}$.

\paragraph{Step 2: $\mathcal{R}(\mathcal{P}_{\mathrm{GCP}})=H^\star$.}
$\mathcal{P}_{\mathrm{GCP}}$ contains the DAG realizing the factorization of
$p^\star$; setting every conditional equal to the corresponding true
conditional gives $p=p^\star$ and hence zero excess KL.

\paragraph{Step 3: $\mathcal{R}(\mathcal{P}_{\mathrm{CBM}})=H^\star+\mathbb{E}[\mathrm{TC}(\mathbf{c}\mid x)]$.}
For $p\in\mathcal{P}_{\mathrm{CBM}}$ the KL chain rule gives
\begin{align*}
&\mathrm{KL}\big(
p^\star(y,\mathbf{c}\mid x)
\,\|\,p(y,\mathbf{c}\mid x)
\big)\\
&\quad=
\mathrm{KL}\big(
p^\star(\mathbf{c}\mid x)
\,\|\,{\textstyle\prod_j} p(c_j\mid x)
\big)\\
&\qquad+
\mathbb{E}_{p^\star(\mathbf{c}\mid x)}
\!\Big[
\mathrm{KL}\big(
p^\star(y\mid x,\mathbf{c})
\,\|\,p(y\mid x,\mathbf{c})
\big)\Big].
\end{align*}
The two terms depend on disjoint parameters and can be minimized
independently. The second vanishes at $p(y\mid x,\mathbf{c})=p^\star(y\mid x,\mathbf{c})$,
which is unconstrained in $\mathcal{P}_{\mathrm{CBM}}$. For the first, writing
$q_j:=p(c_j\mid x)$,
\begin{align*}
&\mathrm{KL}\big(
p^\star(\mathbf{c}\mid x)
\,\|\,{\textstyle\prod_j}q_j
\big)\\
&\quad=
-H\big(p^\star(\mathbf{c}\mid x)\big)
-\sum_j\mathbb{E}_{p^\star}\big[\log q_j(c_j)\big]\\
&\quad=
\mathrm{TC}(\mathbf{c}\mid x)
+\sum_j\mathrm{KL}\big(
p^\star(c_j\mid x)\,\|\,q_j
\big),
\end{align*}
using $\mathrm{TC}(\mathbf{c}\mid x)=\sum_j H(p^\star(c_j\mid x))-H(p^\star(\mathbf{c}\mid x))$.
By Gibbs' inequality this is minimized at $q_j=p^\star(c_j\mid x)$, with
minimum exactly $\mathrm{TC}(\mathbf{c}\mid x)$, i.e.\ the $I$-projection of
$p^\star(\mathbf{c}\mid x)$ onto the product family is the product of its
marginals. Taking expectations over $x$ gives the claim.

\paragraph{Step 4: $\mathcal{R}(\mathcal{P}_{\mathrm{MLP}})=H^\star+\mathbb{E}[\mathrm{TC}(\mathbf{c}\mid x)]+\mathbb{E}[I(y;\mathbf{c}\mid x)]$.}
The same decomposition applies, and the concept term again contributes
$\mathrm{TC}(\mathbf{c}\mid x)$. For the label term, $p(y\mid x)$ may not depend on
$\mathbf{c}$, so we must evaluate
\begin{align*}
&\inf_{q(\cdot)}
\mathbb{E}_{p^\star(\mathbf{c}\mid x)}
\!\Big[
\mathrm{KL}\big(p^\star(y\mid x,\mathbf{c})\,\|\,q\big)
\Big]\\
&\quad=
-H(y\mid x,\mathbf{c})
-\sup_{q}\sum_y p^\star(y\mid x)\log q(y)\\
&\quad=I(y;\mathbf{c}\mid x),
\end{align*}
where the supremum is attained at $q=p^\star(y\mid x)$ by Gibbs'
inequality and we used
$\mathbb{E}_{p^\star(\mathbf{c}\mid x)}[p^\star(y\mid x,\mathbf{c})]=p^\star(y\mid x)$.

\paragraph{Step 5: strictness.}
$\mathrm{TC}(\mathbf{c}\mid x)\ge0$ with equality iff $p^\star(\mathbf{c}\mid x)$ is a
product, so if $c_i\not\!\perp\!\!\!\perp c_j\mid x$ for some $i\neq j$ on a set of
$x$ of positive probability then
$\mathbb{E}[\mathrm{TC}(\mathbf{c}\mid x)]>0$ and
$\mathcal{R}(\mathcal{P}_{\mathrm{GCP}})<\mathcal{R}(\mathcal{P}_{\mathrm{CBM}})$.
Likewise $I(y;\mathbf{c}\mid x)\ge0$ with equality iff
$y\perp\mathbf{c}\mid x$, giving the second strict inequality under the
stated condition. \qed

\subsection{Proof of Theorem~\ref{thm:gcp_conv}}

\label{sec:proof_gcp_conv}

\textbf{Step 0: separability and smoothness.}
Under Assumption~\ref{asm:teacher_forcing} each $F_v$ depends on
$\theta_v$ only, so $F(\theta)=\sum_{v\in V}\lambda_v F_v(\theta_v)$ has
block-diagonal Hessian and
$\nabla F(\theta)=\big(\lambda_v\nabla F_v(\theta_v)\big)_{v\in V}$.
Consequently $F$ is $L$-smooth with $L=\max_{v}\lambda_v L_v$, and its
minimizer is blockwise, $\theta^\star=(\theta_v^\star)_{v\in V}$ with
$\theta_v^\star\in\arg\min F_v$, so that
$F(\theta)-F(\theta^\star)=\sum_v \lambda_v\big(F_v(\theta_v)-F_v(\theta_v^\star)\big)$
with every summand nonnegative.

\textbf{Step 1: PL for the sum objective.}
Using separability and then the node-level PL inequalities
$\|\nabla F_v(\theta_v)\|^2\ge 2\mu_v\big(F_v(\theta_v)-F_v(\theta_v^\star)\big)$,
\[
\|\nabla F(\theta)\|^2
=
\sum\nolimits_{v\in V}\lambda_v^2\,\|\nabla F_v(\theta_v)\|^2
\]
\[
\ge
2\sum\nolimits_{v\in V}\lambda_v^2\,\mu_v\big(F_v(\theta_v)-F_v(\theta_v^\star)\big)
\]
\[
=
2\sum\nolimits_{v\in V}\big(\lambda_v\mu_v\big)\,\lambda_v\big(F_v(\theta_v)-F_v(\theta_v^\star)\big)
\]
\[
\ge
2\Big(\min_{v\in V}\lambda_v\mu_v\Big)
\sum\nolimits_{v\in V}\lambda_v\big(F_v(\theta_v)-F_v(\theta_v^\star)\big),
\]
where the last step uses $\lambda_v>0$ and the nonnegativity of each
summand. With $\mu_{\mathrm{GCP}}=\min_{v}\lambda_v\mu_v$ this is exactly
\[
\tfrac{1}{2}\|\nabla F(\theta)\|^2
\ge
\mu_{\mathrm{GCP}}\big(F(\theta)-F(\theta^\star)\big)
\]
on the intersection of the node-level neighbourhoods, i.e.\ $F$ satisfies
a local PL inequality with constant $\mu_{\mathrm{GCP}}$. For uniform
weights $\lambda_v\equiv1$ this reduces to $\min_v\mu_v$.

\textbf{Step 2: smooth ${}+{}$ PL $\Rightarrow$ linear convergence.}
The descent lemma for $L$-smooth $F$ with step size $1/L$ gives
$F(\theta_{t+1})\le F(\theta_t)-\tfrac{1}{2L}\|\nabla F(\theta_t)\|^2$;
substituting the PL inequality of Step~1 yields
\[
F(\theta_{t+1})-F(\theta^\star)
\le
\Big(1-\frac{\mu_{\mathrm{GCP}}}{L}\Big)\big(F(\theta_t)-F(\theta^\star)\big),
\]
and iterating gives the stated rate in function value. If $F$ further
satisfies the quadratic growth condition
$F(\theta)-F(\theta^\star)\ge\tfrac{\mu_{\mathrm{GCP}}}{2}\|\theta-\theta^\star\|^2$
on the same neighbourhood, the
contraction transfers to parameter distance. \qed

\medskip
\noindent\textbf{Proof of Proposition~\ref{prop:mu_order}.}
By Assumption~\ref{asm:curvature}, $\mu_v^{\mathrm{GCP}}\ge\mu_v^{\mathrm{CBM}}\ge\mu_v^{\mathrm{MLP}}$
for every $v$, hence
$\min_v\lambda_v\mu_v^{\mathrm{GCP}}\ge\min_v\lambda_v\mu_v^{\mathrm{CBM}}\ge\min_v\lambda_v\mu_v^{\mathrm{MLP}}$.
If both comparisons are strict for every node, finiteness of $V$ and
positivity of the weights give positive gaps
$\delta_1=\min_v\lambda_v(\mu_v^{\mathrm{GCP}}-\mu_v^{\mathrm{CBM}})$
and
$\delta_2=\min_v\lambda_v(\mu_v^{\mathrm{CBM}}-\mu_v^{\mathrm{MLP}})$.
Thus
$\mu_{\mathrm{GCP}}\ge\mu_{\mathrm{CBM}}+\delta_1>
\mu_{\mathrm{CBM}}$ and
$\mu_{\mathrm{CBM}}\ge\mu_{\mathrm{MLP}}+\delta_2>
\mu_{\mathrm{MLP}}$. The rate comparison then follows from
Theorem~\ref{thm:gcp_conv}. \qed

\medskip
\noindent\textbf{On verifying Assumption~\ref{asm:curvature}.}
The assumption is stated rather than derived, and we do not claim a
general proof: for a categorical node the Fisher information in logit
space is $\mathrm{diag}(p)-pp^{\!\top}$, which \emph{decreases} as the
conditional becomes more concentrated, so a naive
``conditioning ${}\Rightarrow{}$ sharper conditional ${}\Rightarrow{}$
larger curvature'' argument is not valid. The plausible mechanism is
instead a reduction in the effective dimension of each node subproblem
under conditioning, and we therefore treat
Assumption~\ref{asm:curvature} as an empirical hypothesis, estimating
node-level PL constants by
$\widehat{\mu}_v=\min_{t}\|\nabla F_v(\theta_v^{(t)})\|^2/\big(2(F_v(\theta_v^{(t)})-\widehat{F}_v^\star)\big)$
along the training trajectory.

To make the comparison well defined at the node level, all three
architectures are trained on the same node set $V$ and differ only in
what each node conditions on: in GCP the predictor $f_v$ reads $x$
together with the realized parent concepts $\mathbf{h}_{\mathrm{pa}(v)}$;
in the flat CBM it reads $x$ alone; and in the MLP variant the concept
predictors share a single trunk, so $\theta_v$ is the corresponding
read-out block rather than an independent module. We estimate
$\widehat{\mu}_v$ for each node along the training trajectory, taking
$\widehat{F}_v^\star$ to be the lowest value that node's objective
reaches. Because the estimator minimizes over $t$, every entry is a
conservative lower bound on the local curvature.

\begin{table}[ht]
\centering
\small
\setlength{\tabcolsep}{4pt}
\caption{\textbf{Estimated node-level PL constants} (all values
$\times 10^{-2}$). $\widehat{\mu}_{\min}=\min_v\widehat{\mu}_v$ is the
quantity that controls the rate in Theorem~\ref{thm:gcp_conv} under
uniform $\lambda_v$; $v^\star$ is the node attaining it.}
\label{tab:pl_constants}
\resizebox{\columnwidth}{!}{%
\begin{tabular}{@{}lccl@{}}
\toprule
Model & $\widehat{\mu}_{\min}$ & median $\widehat{\mu}_v$ & $v^\star$ \\
\midrule
\multicolumn{4}{@{}l}{\emph{MNLI} ($|V|=16$)} \\
MLP       & 0.63 & 1.9 & negation scope \\
flat CBM  & 1.10 & 3.2 & negation scope \\
GCP       & \textbf{1.80} & \textbf{4.6} & negation scope \\
\midrule
\multicolumn{4}{@{}l}{\emph{MIMIC-III} ($|V|=18$)} \\
MLP       & 0.51 & 1.5 & lab abnormality \\
flat CBM  & 0.92 & 2.7 & treatment escalation \\
GCP       & \textbf{1.40} & \textbf{3.9} & treatment escalation \\
\bottomrule
\end{tabular}%
}
\end{table}

Three observations follow. First, the ordering asserted by
Assumption~\ref{asm:curvature} is respected on both datasets, at the
median as well as at the minimum, which is the form in which
Proposition~\ref{prop:mu_order} actually uses it. Second, the gap is
$\approx\!2.9\times$ on MNLI and $\approx\!2.7\times$ on
MIMIC-III between GCP and the MLP; since $\widehat{\mu}_{\min}$ is
exactly the quantity entering the contraction factor
$1-\mu_{\mathrm{GCP}}/L$, this is the sense in which conditioning buys a
faster rate, and it is a statement about curvature at matched $L$ rather
than a measured wall-clock speedup. Third, the bottleneck node is
stable on MNLI but not on MIMIC-III: the MLP is limited by
\emph{lab abnormality}, whereas both concept-based models are limited by
\emph{treatment escalation}. This illustrates why strictness must be
checked at the level of the minima: the minimum can increase even when
the identity of the minimizing node changes, while an improvement at
only a non-minimizing node would not suffice.
Consistent with our framing of Assumption~\ref{asm:curvature} as an
empirical hypothesis, these estimates are evidence for the ordering on
the tasks we study, not a proof of it.

\subsection{Proof of Theorem~\ref{thm:optimal}: Optimality}

\label{sec:proof_optimal}

Let $\{\widehat{\Delta}_{(1)}\ge \widehat{\Delta}_{(2)}\ge \cdots \ge \widehat{\Delta}_{(|V|)}\}$
be the sorted scores, and let $S^\star$ be the set of indices of the top-$b$ scores.
For any feasible $S$ with $|S|=b$, if $S\neq S^\star$, then there exists
$i\in S\setminus S^\star$ and $j\in S^\star\setminus S$ such that
$\widehat{\Delta}_j \ge \widehat{\Delta}_i$.
Swapping $i$ out and $j$ in does not decrease the objective:
\[
\sum\nolimits_{\ell\in (S\setminus\{i\})\cup\{j\}} \widehat{\Delta}_\ell
=
\sum\nolimits_{\ell\in S} \widehat{\Delta}_\ell - \widehat{\Delta}_i + \widehat{\Delta}_j
\]
\[
\ge
\sum\nolimits_{\ell\in S} \widehat{\Delta}_\ell.
\]
By repeating this exchange argument finitely many times, any feasible $S$ can be
transformed into $S^\star$ without decreasing the objective. Hence $S^\star$ is
globally optimal.

\subsection{Proof of Theorem~\ref{thm:optimal}: Time Complexity}

\label{sec:proof_timecomplexity}

In the worst case, Algorithm~\ref{alg:submodule_retrain} evaluates (or replays)
counterfactual losses for each node and for each example. Let $|V|$ be the number
of nodes. For a fixed example, computing node-wise contributions can require
propagating information across the graph; under dense connectivity or full
pairwise interactions this costs $\mathcal{O}(|V|^2)$ per example. Aggregating over
$|\mathcal{D}|$ examples yields $\mathcal{O}(|\mathcal{D}||V|^2)$ total time in the worst case. Lower-order
terms (e.g., sorting $|V|$ scores, $\mathcal{O}(|V|\log |V|)$) are dominated.

\section{Additional Experimental Results}
\label{sec:additional_results}

\subsection{Token--Accuracy Trade-Off of Targeted Retraining}
\label{sec:token_accuracy}

\begin{figure*}[t!]
\small
\centering
\includegraphics[width=0.9\textwidth]{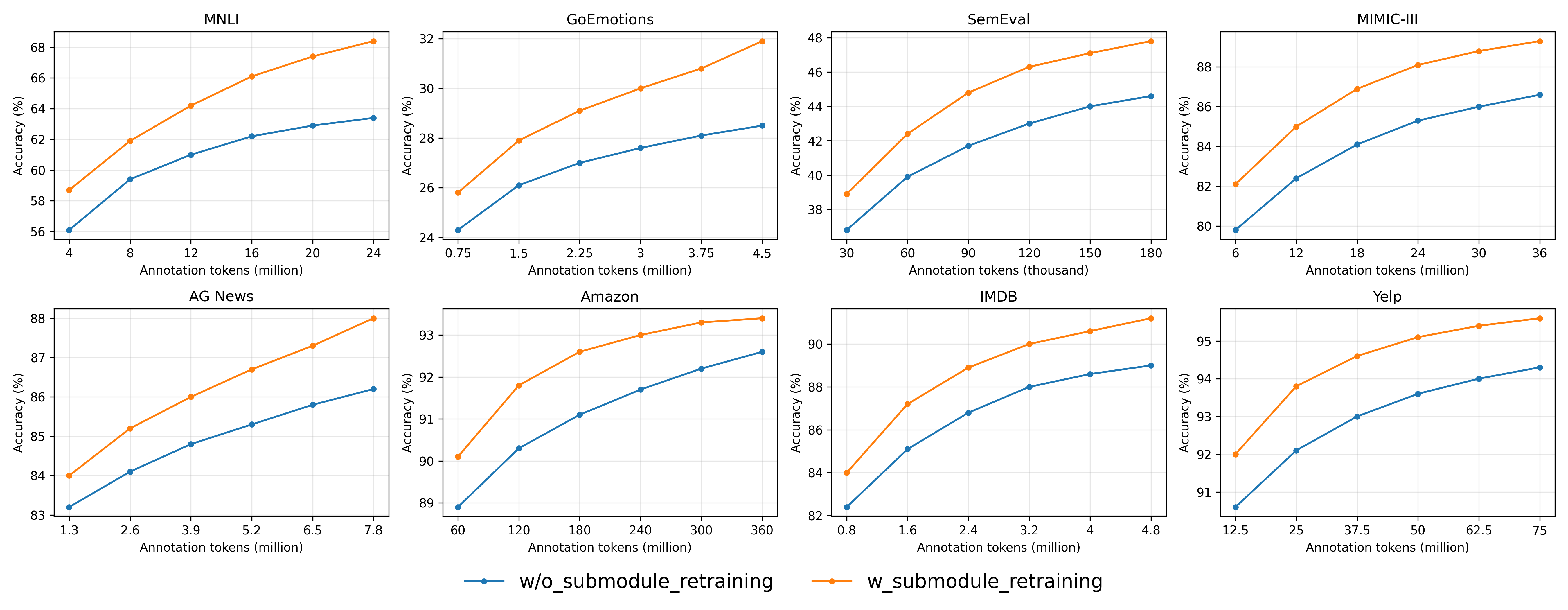}
\caption{\textbf{Ablation study of targeted sub-module retraining.}
The y-axis denotes classification accuracy and the x-axis denotes
teacher-token usage (millions; thousands for SemEval). Budgets are set per
dataset and span roughly 8--50\% of the tokens needed to label its full
training set with labels only (cf.\ the 5--50\% range of
Figure~\ref{fig:draw_trade_off}).}
\label{fig:trade_off_token}
\end{figure*}

Figure~\ref{fig:trade_off_token} shows that sub-module retraining
consistently improves the token--accuracy trade-off across all annotation
budgets. Because label-only annotation cost scales with training-set size and
text length, each dataset receives six evenly spaced token budgets that end
near 50\% of its label-only cost, from 30--180K tokens on SemEval to
60--360M on Amazon. On every dataset, the retraining curve dominates the
non-retraining curve. The largest and most persistent gains occur on MNLI,
SemEval, and MIMIC-III, typically 2--5 points. For example, at the largest
budget, MNLI improves from 63.4 to 68.4 (24M tokens) and SemEval from 44.6 to
47.8 (180K tokens). Clear
improvements also appear on the high-accuracy sentiment and topic tasks
(Yelp, Amazon, and AG News), typically by 1--2 points. Retraining also
improves token efficiency, often matching or exceeding no-retraining
performance at substantially lower budgets.

\subsection{Computational Cost and Runtime}
\label{sec:runtime}

\begin{figure}[ht]
\centerline{\includegraphics[width=0.7\columnwidth]{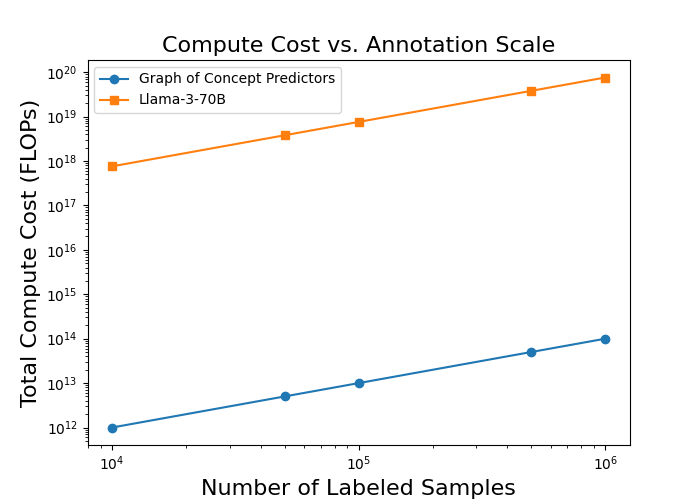}}
    \vspace{-2mm}
    \caption{\textbf{Compute cost versus annotation scale on a logarithmic
    axis.} Comparison between the LLaMA-3-70B teacher and GCP.}
\label{fig:compute_cost_vs_annotation_scale}
\end{figure}

Table~\ref{tab:runtime} reports the wall-clock cost of one active-learning
round, including acquisition and retraining, on a single H200 GPU at a
20\% budget. GCP takes 1.3--1.6$\times$ BADGE's per-round time because of
counterfactual sub-module retraining, but remains tractable across all four
datasets. Teacher-inference cost at larger annotation scales is reported
separately in Figure~\ref{fig:compute_cost_vs_annotation_scale}.

\begin{table}[ht]
\centering
\small
\setlength{\tabcolsep}{5pt}
\caption{\textbf{Per-round runtime (seconds) at 20\% budget.}}
\label{tab:runtime}
\begin{tabular}{lcccc}
\toprule
Method & MNLI & SemEval & MIMIC-III & Yelp \\
\midrule
Entropy   & 38  & 21  & 64  & 73 \\
CoreSet   & 62  & 34  & 102 & 121 \\
BADGE     & 94  & 51  & 158 & 187 \\
CAL       & 112 & 67  & 174 & 213 \\
GCP       & 137 & 82  & 211 & 248 \\
\bottomrule
\end{tabular}
\end{table}

\section{Concept Graph Construction}
\label{sec:concept_graph}

This appendix gives the full specification of the construction procedure
summarized in Section~\ref{sec:graph_construction}.
Algorithm~\ref{alg:graph_construction} states the procedure in full, and
Table~\ref{tab:graph_stats} reports the resulting graph statistics for each
dataset.

\paragraph{Division of labour between the teacher and the data.}
It is important to distinguish the parts of the final graph supplied by the
teacher LLM, by a fixed name-similarity rule, and by annotated data, because
these sources have different failure modes:

\begin{itemize}[leftmargin=1.2em,itemsep=0.15em,topsep=0.25em]
\item \textbf{Decided by the LLM:} the concept vocabulary (which
intermediate quantities exist and what they mean), and the causal
ordering $\prec$ used to orient edges. A concept the teacher never
proposes cannot enter the graph.
\item \textbf{Decided by a fixed name-similarity rule:} which near-duplicate
concepts are removed. Each concept $c$ is embedded as
$\mathrm{SBERT}(\mathrm{name}(c))$~\cite{reimers2019sentencebertsentenceembeddingsusing},
where $\mathrm{name}(c)$ is the concept name written by the teacher; its
short definition is not encoded. If two embeddings have cosine similarity
above $\rho$,
the later concept in $\prec$ is dropped together with its candidate edges.
This step uses no annotated data.
\item \textbf{Decided by annotated data:} which remaining concepts are
kept (entropy filter) and, critically, \emph{which candidate dependencies
become edges} (the conditional mutual information test). A dependency the
teacher asserts but that carries no measurable conditional information is
discarded.
\end{itemize}

\noindent
The graph is therefore not a raw zero-shot LLM output. The teacher proposes
the vocabulary and candidate structure, the name-similarity rule removes
redundant concepts, and statistical tests on the annotations of
$\mathcal{X}_0$ decide which of
the remaining concepts and dependencies survive. In particular, unsupported
or hallucinated dependencies are filtered before they can influence
acquisition or retraining.

\paragraph{The initial sample is training data.}
The sample $\mathcal{X}_0$ is drawn from the unlabeled pool $\mathcal{U}$ and is not
held out. The teacher's annotations of $\mathcal{X}_0$ select $V$ and $E$
and then initialize $\mathcal{D}$ (Algorithm~\ref{alg:gcp}, Line~1), so the
student is first trained on the same annotations that chose its graph. The
entropy and conditional mutual information tests ask whether a concept or a
dependency is supported by the teacher's annotations, not how well the
student generalizes. The test split plays no part in construction, so
reported accuracies are measured on data that did not influence the graph.

\paragraph{Acyclicity by construction.}
The ordering $\prec$ is the position of each concept in the teacher's list,
so it is a strict total order. Stage~1 keeps only candidate edges
$(c_i,c_j)$ with $c_i\prec c_j$ and discards any citation of a later
concept or of the concept itself. Every edge in $E$ therefore points from
an earlier to a later concept, positions strictly increase along any
directed path, and $G$ cannot contain a cycle. The later stages only remove
nodes and edges, which preserves this property, so no post-hoc pruning is
needed.

\begin{algorithm}[ht]
\caption{Concept graph construction (run once per task)}
\label{alg:graph_construction}
\begin{algorithmic}[1]
\STATE {\bfseries Input:} task definition $T$, label set $\mathcal{Y}$,
representative unlabeled examples $\mathcal{X}_{\mathrm{ex}}$, initial
sample $\mathcal{X}_0\subset\mathcal{U}$,
teacher $\mathcal{T}$, thresholds $\tau=0.15$, $\rho=0.85$, $\eta=0.95$
\STATE {\bfseries Output:} concept DAG $G=(V,E)$
\STATE \textit{// Stage 1: elicitation}
\STATE $(C,\prec,E_{\mathrm{cand}}) \leftarrow \mathcal{T}(T,\mathcal{Y},\mathcal{X}_{\mathrm{ex}})$
\hfill {\small// $30$--$100$ ordered concepts with cited dependencies}
\STATE $E_{\mathrm{cand}} \leftarrow \{(c_i,c_j)\in E_{\mathrm{cand}} : c_i \prec c_j\}$
\hfill {\small// discard citations of later concepts}
\STATE \textit{// Stage 2a: deduplication by name (no data)}
\STATE $\mathbf{e}(c) \leftarrow \mathrm{SBERT}(\mathrm{name}(c))$, $\forall c\in C$
\FOR{each pair $c_i,c_j\in C$ with $c_i \prec c_j$}
  \IF{$\cos(\mathbf{e}(c_i),\mathbf{e}(c_j)) > \rho$}
     \STATE $C \leftarrow C\setminus\{c_j\}$ \hfill {\small// remove later near-duplicate}
  \ENDIF
\ENDFOR
\STATE \textit{// Stage 2b: entropy filter on $\mathcal{X}_0$}
\STATE annotate $\mathcal{X}_0$ with $\mathcal{T}$ for all $c\in C$
\STATE $V \leftarrow \{c\in C : H(c)/\log 2 \le \eta\}$ \hfill {\small// drop near-random concepts}
\STATE \textit{// Stage 3: edge induction}
\STATE $E \leftarrow \varnothing$
\FOR{each $(c_i,c_j)\in E_{\mathrm{cand}}$ with $c_i,c_j\in V$}
  \IF{$\widehat{I}(c_i;c_j\mid x) / \min\{H(c_i),H(c_j)\} > \tau$}
     \STATE $E \leftarrow E\cup\{(c_i \rightarrow c_j)\}$ \hfill {\small// follows $\prec$}
  \ENDIF
\ENDFOR
\STATE \textbf{return} $G=(V,E)$ \hfill {\small// structure fixed for all AL rounds}
\end{algorithmic}
\end{algorithm}

\paragraph{Cost.}
Construction uses one elicitation call of about 5K tokens and one teacher
annotation pass over $\mathcal{X}_0$ (Stage~2b). The tokens of both steps
are charged once to GCP's token budget $N$, the same budget every method
receives.

\paragraph{Graph statistics.}
Table~\ref{tab:graph_stats} reports, for each benchmark, the number of
candidate concepts elicited in Stage~1, the node count after pruning, the
number of candidate dependencies, and the number of edges surviving the
conditional mutual information test. The gap between
$|E_{\mathrm{cand}}|$ and $|E|$ is the fraction of teacher-proposed
structure rejected by the annotated data, which ranges from $28$\% to $40$\% of the
candidate edges across benchmarks. Graph size is task-dependent rather
than fixed manually. The topic and sentiment benchmarks retain $5$--$8$
concepts, whereas the multi-factor benchmarks retain $14$--$19$ and preserve
more conditional dependence among them. This pattern is consistent with
Theorem~\ref{thm:gcp_risk}: graph structure is most useful when the retained
concepts exhibit nontrivial conditional total correlation.

\begin{table}[ht]
\centering
\small
\setlength{\tabcolsep}{4pt}
\caption{\textbf{Concept graph statistics per dataset.} $|C|$: candidate
concepts elicited in Stage~1; $|V|$: nodes retained after pruning;
$|E_{\mathrm{cand}}|$: teacher-proposed dependencies whose endpoints both
survive pruning, i.e.\ those actually submitted to the CMI test; $|E|$:
edges retained after that test. Node count and edge density both track
reasoning depth rather than input length.}
\label{tab:graph_stats}
\begin{tabular}{@{}lcccc@{}}
\toprule
Dataset & $|C|$ & $|V|$ & $|E_{\mathrm{cand}}|$ & $|E|$ \\
\midrule
\multicolumn{5}{@{}l}{\emph{Topic / sentiment}} \\
AG News    & 32 & 6  & 11 & 7  \\
Amazon     & 35 & 8  & 15 & 9  \\
IMDB       & 31 & 5  & 9  & 6  \\
Yelp       & 33 & 7  & 13 & 9  \\
\midrule
\multicolumn{5}{@{}l}{\emph{Multi-factor reasoning}} \\
MNLI       & 41 & 16 & 34 & 23 \\
GoEmotions & 48 & 19 & 41 & 26 \\
SemEval    & 38 & 14 & 29 & 21 \\
MIMIC-III  & 45 & 18 & 38 & 26 \\
\bottomrule
\end{tabular}
\end{table}

Table~\ref{tab:concept_examples} lists three representative concepts per
dataset, illustrating the kind of mid-level reasoning units captured by
GCP.

\begin{table}[ht]
\centering
\small
\setlength{\tabcolsep}{4pt}
\caption{\textbf{Representative concepts induced for each dataset.}}
\label{tab:concept_examples}
\begin{tabular}{@{}lp{0.66\columnwidth}@{}}
\toprule
Dataset & Example concepts \\
\midrule
AG News        & \emph{geopolitical entity, market terminology, scientific jargon} \\
Amazon         & \emph{product defect mention, comparative phrasing, sentiment polarity} \\
IMDB           & \emph{cast praise, plot critique, recommendation cue} \\
Yelp           & \emph{service quality, food descriptor, price sentiment} \\
MNLI           & \emph{lexical overlap, negation scope, temporal mismatch} \\
GoEmotions     & \emph{intensifier, second-person address, gratitude cue} \\
SemEval        & \emph{target relevance, sarcasm cue, authority appeal} \\
MIMIC-III      & \emph{comorbidity mention, lab abnormality, treatment escalation} \\
\bottomrule
\end{tabular}
\end{table}

\section{Annotation Prompt Templates}
\label{sec:prompts}

The pipeline uses three types of teacher call, and we provide the
corresponding LLaMA-3-70B template for each below. Full per-dataset templates are released with
the code.

\subsection{Concept Graph Elicitation}
\label{sec:prompt_elicitation}

This prompt is issued once per task. It produces the candidate concept
vocabulary, dependency order, and candidate edge set consumed by Stage~1 of
Algorithm~\ref{alg:graph_construction}. Requiring concepts to be listed in
dependency order yields the causal ordering $\prec$ used to orient edges.

\begin{quote}\small
\texttt{You are designing an interpretable classifier for the following
task.\\
Task: \{task\_definition\}\\
Label set: \{labels\}\\
Example inputs: \{representative\_examples\}\\
List 30-100 intermediate concepts that a careful human annotator would
check when assigning a label for this task. Each concept must be a
binary property of the input.\\
Order the list so that a concept appears only after every concept it
depends on. For each concept, cite by index the earlier concepts whose
value would change your judgement of it; write NONE if there are none.\\
Return one concept per line in the format:\\
index | concept name | short definition | depends\_on: [indices]}
\end{quote}

\subsection{Joint Label and Concept Annotation}
\label{sec:prompt_annotation}

Each selected instance is annotated in a single teacher call that returns
both the final label and the values of all concept nodes in $V$. Combining
the two outputs keeps the per-instance overhead close to label-only
annotation because the label decision and concept flags share one pass over
the input.

\begin{quote}\small
\texttt{You are a precise classifier and concept annotator.\\
Text: \{text\}\\
Label set: \{labels\}\\
Concepts: \{concept\_list\}\\
Return exactly one label for a single-label task, or every applicable
label for a multi-label task. Then, for each concept in the order given,
decide whether it is present (1) or absent (0) in the text. Use the
parent concept values you have already assigned as context when judging
a later concept.\\
Return only:\\
\{"label": <label or label list>, "concept": [<0/1>, ...]\}}
\end{quote}

\subsection{Parent--Child Concept Generation}
\label{sec:prompt_regeneration}

During sub-module retraining (Section~\ref{sec:submodule}), the teacher is
queried once per selected bottleneck node to generate additional
parent--child concept pairs as focused supervision for that predictor.

\begin{quote}\small
\texttt{You are generating additional training examples for a
single step of a concept-based classifier.\\
Task: \{task\_definition\}\\
Target concept: \{child\_name\} -- \{child\_definition\}\\
Parent concepts, in dependency order:\\
\{parent\_name\} -- \{parent\_definition\}\\
Write \{K\} short passages in the style of the task above. For each
passage, assign a value (1 = present, 0 = absent) to every parent
concept and to the target concept. The target value must be consistent
with the parent values you assign and with the definitions given
above.\\
Use both values of the target concept about equally often, vary the
parent configurations across passages, and do not reuse phrasing. Do
not state or imply the final task label.\\
Return one example per line in the format:\\
\{"text": <passage>, "parents": [<0/1>, ...], "target": <0/1>\}}
\end{quote}

Two safeguards limit how the generated data can affect the results. First,
the prompt forbids stating or implying the final task label, and each
generated pair contains concept values but no task label. The pairs therefore
enter only the node-local objective for the selected predictor $f_i$: the
label predictor $f_y$ is never trained on synthetic data, and concept annotations
for real instances are never overwritten. Second, only the $b$ selected predictors
are updated, each starting from its current weights and trained on the real
annotations in $\mathcal{D}$ together with its generated pairs (Line~9 of
Algorithm~\ref{alg:gcp}), and all other predictors stay frozen. We generate $K=200$ pairs per
selected node and remove exact-duplicate passages before retraining.

\end{document}